\def\year{2022}\relax
\documentclass[letterpaper]{article} 
\usepackage{aaai22}  
\usepackage{times}  
\usepackage{helvet}  
\usepackage{courier}  
\usepackage[hyphens]{url}  
\usepackage{graphicx} 
\usepackage{natbib}  
\usepackage{caption} 
\DeclareCaptionStyle{ruled}{labelfont=normalfont,labelsep=colon,strut=off} 
\usepackage{algorithm}
\usepackage{algorithmic}

\usepackage{url}
\usepackage{graphicx}
\usepackage{cite}
\usepackage{amsmath,amssymb,amsfonts}
\usepackage{textcomp}
\usepackage{subfigure}
\usepackage{caption}
\usepackage{xcolor}
\usepackage{booktabs}
\usepackage{arydshln}
\usepackage{bbding}
\usepackage{multirow}
\usepackage{diagbox}
\usepackage{ntheorem}
\usepackage{makecell}
\usepackage{colortbl}
\usepackage{pifont}
\usepackage{tablefootnote}

\usepackage[pagebackref=true,breaklinks=true,colorlinks=true,letterpaper=true,bookmarks=false,linkcolor={red!50!black},urlcolor={darkgray!80!black},citecolor={green!50!black}]{hyperref} 
\usepackage{xspace} 
\makeatletter
\DeclareRobustCommand\onedot{\futurelet\@let@token\@onedot}
\def\@onedot{\ifx\@let@token.\else.\null\fi\xspace}

\def\eg{\emph{e.g}\onedot} 
\def\ie{\emph{i.e}\onedot} 
 
\def\etc{\emph{etc}\onedot} 
 
\def\etal{\emph{et al}\onedot}
\makeatother

\usepackage[switch]{lineno} 
\usepackage{siunitx}
\newcommand{\figref}[1]{Fig.~\ref{#1}}

\newcommand{\tabref}[1]{Table~\ref{#1}}

\newtheorem{theorem}{Theorem}

\definecolor{seagreen}{RGB}{84,255,159}
\definecolor{SpringGreen}{RGB}{0,139,69}
\usepackage[font=small,labelfont=bf]{caption}

\usepackage{booktabs}
\usepackage{amsmath}
\usepackage{multirow}

\usepackage{newfloat}
\usepackage{listings}
\floatstyle{ruled}
\newfloat{listing}{tb}{lst}{}
\floatname{listing}{Listing}
\title{End-to-end Learning the Partial Permutation Matrix \\ for Robust 3D Point Cloud Registration}
\author {
    Zhiyuan Zhang\textsuperscript{\rm 1},
    Jiadai Sun\textsuperscript{\rm 1},
    Yuchao Dai\textsuperscript{\rm 1}\thanks{Yuchao Dai is the corresponding author.},
    Dingfu Zhou\textsuperscript{\rm 2},
    Xibin Song\textsuperscript{\rm 2},
    Mingyi He\textsuperscript{\rm 1}
}
\affiliations {
    \textsuperscript{\rm 1} Northwestern Polytechnical University \quad
    \textsuperscript{\rm 2} Baidu Research. \\
    \{zhangzhiyuan,sunjiadai\}@mail.nwpu.edu.cn, daiyuchao@gmail.com,\\ \{zhoudingfu,songxibin\}@baidu.com, myhe@nwpu.edu.cn
}

\usepackage{bibentry}

\begin{document}

\maketitle

\begin{abstract}
Even though considerable progress has been made in deep learning-based 3D point cloud processing, how to obtain accurate correspondences for robust registration remains a major challenge because existing hard assignment methods cannot deal with outliers naturally. Alternatively, the soft matching-based methods have been proposed to learn the matching probability rather than hard assignment. However, in this paper, we prove that these methods have an inherent ambiguity causing many deceptive correspondences.
To address the above challenges, we propose to learn a partial permutation matching matrix, which does not assign corresponding points to outliers, and implements hard assignment to prevent ambiguity. 
However, this proposal poses two new problems, \ie existing hard assignment algorithms can only solve a full rank permutation matrix rather than a partial permutation matrix, and this desired matrix is defined in the discrete space, which is non-differentiable.
In response, we design a dedicated soft-to-hard (S2H) matching procedure within the registration pipeline consisting of two steps: solving the soft matching matrix (S-step) and projecting this soft matrix to the partial permutation matrix (H-step).
Specifically, we augment the profit matrix before the hard assignment to solve an augmented permutation matrix, which is cropped to achieve the final partial permutation matrix. Moreover, to guarantee end-to-end learning, we supervise the learned partial permutation matrix but propagate the gradient to the soft matrix instead.
Our S2H matching procedure can be easily integrated with existing registration frameworks, which has been verified in representative frameworks including DCP, RPMNet, and DGR. Extensive experiments have validated our method, which creates a new state-of-the-art performance for robust 3D point cloud registration. \textit{The code will be made public.}
\end{abstract}

\section{Introduction}\label{sec:introduction}

3D point cloud registration is a well-known task in 3D vision with wide applications including object pose estimation \cite{ming_segicp_iros_2017}, 3D reconstruction \cite{deschaud_imls_icra_2018}, simultaneous localization and mapping  \cite{shiratori_largeScale_3dv_2015,ding_deepmapping_cvpr_2019}, \etc. 
Although the increasingly prosperous deep learning technique has achieved great success in point cloud registration \cite{lu_deepvcp_iccv_2019}, how to obtain accurate correspondences for robust registration remains a stubborn problem, which can be formulated as solving a matching matrix to relate the input two point clouds. And each entry indicates the point pair is a correspondence or not.

For ideal consistent point clouds, where the inputs are exactly the same except for the pose, \ie each point can find a corresponding point in the other point cloud, the correspondences are built by searching a permutation matching matrix, which implements the one-to-one matching principle. 
However, in practical applications, input point clouds are usually not consistent due to the outliers (\ie the points without corresponding points).
To handle the outliers, a widely adopted strategy is to select reliable correspondences after the initial matching \cite{choy_dgr_cvpr_2020,probst_consensusMax_cvpr_2019,pais_3dregnet_cvpr_2020,PPFNet_Deng_2018_CVPR,PPFFolding_Deng_2018_ECCV,PointDSC_Bai_2021_CVPR}. 
Nonetheless, this remedial operation is complicated. Alternatively, we propose to handle the outliers in the matching stage synchronously. 
In this case, the desired matching matrix is turned to a {\underline{p}artial \underline{p}ermutation \underline{m}atrix} (notated as {PPM}) formulated by a binary matrix, where the sum of row or column corresponding to inlier/outlier is one/zero. 
PPM embeds two important principles: one-to-one matching and outliers pruning. 
Unfortunately, existing hard assignment algorithms are not competent to directly solve this PPM since they cannot distinguish inliers and outliers, and they will solve a full rank permutation matrix assigning corresponding points to outliers incorrectly.

\newcommand{\tabincell}[2]{\begin{tabular}{@{}#1@{}}#2\end{tabular}} 
\begin{table*}[!t]
	\centering
	\setlength\tabcolsep{20pt}
	\resizebox{0.85\linewidth}{!}{
    	\begin{tabular}{ccc|cc}
    		\toprule
    		\textbf{Methods} &DCP (Soft matching)  & RPMNet (Soft matching) & PRNet (Hard matching) & \textbf{Ours} (Hard matching) \\
    		\midrule
    		\tabincell{c}{\textbf{Matrix}\\\textbf{Constraints}}
    		&  \tabincell{c}{$ m_{ij} \in [0, 1]$   \\ $\sum_{j}^{N_\mathcal{Y}} m_{ij} = 1$ \\ } 
    		&  \tabincell{c}{$ m_{ij} \in [0, 1]$   \\ $\sum_{i}^{N_\mathcal{X}} m_{ij} \in [0,1]$ \\$\sum_{j}^{N_\mathcal{Y}} m_{ij} \in [0,1]$ \\ }
    		&  \tabincell{c}{$ m_{ij} \in \{0, 1\}$ \\ $\sum_{j}^{N_\mathcal{Y}} m_{ij}=1$ \\ }
    		&  \tabincell{c}{$ m_{ij} \in \{0, 1\}$ \\ $\sum_{i}^{N_\mathcal{X}} m_{ij} \in \{0,1\}$ \\ $\sum_{j}^{N_\mathcal{Y}} m_{ij}\in \{0,1\}$} \\ 
    		
    		\tabincell{c}{\textbf{Matching}\\\textbf{Results}}
    		& \tabincell{c}{\includegraphics[width=1.0in]{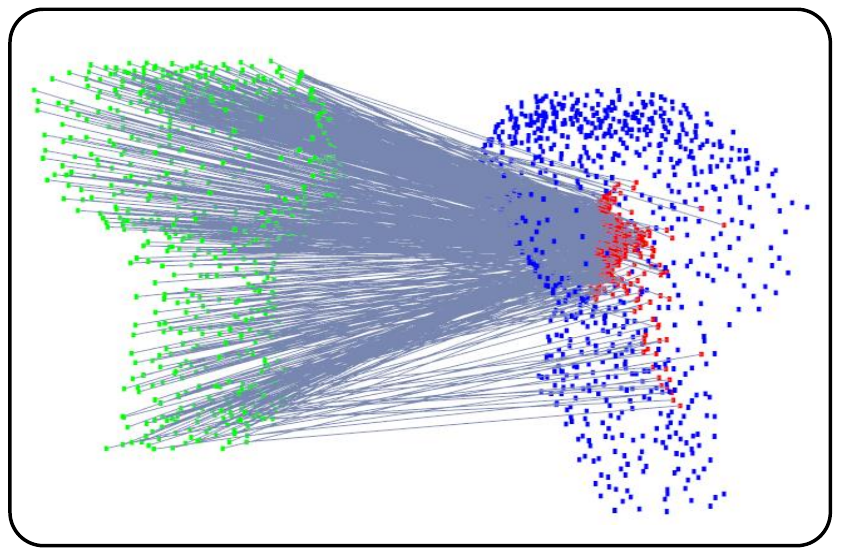}}
    		& \tabincell{c}{\includegraphics[width=1.0in]{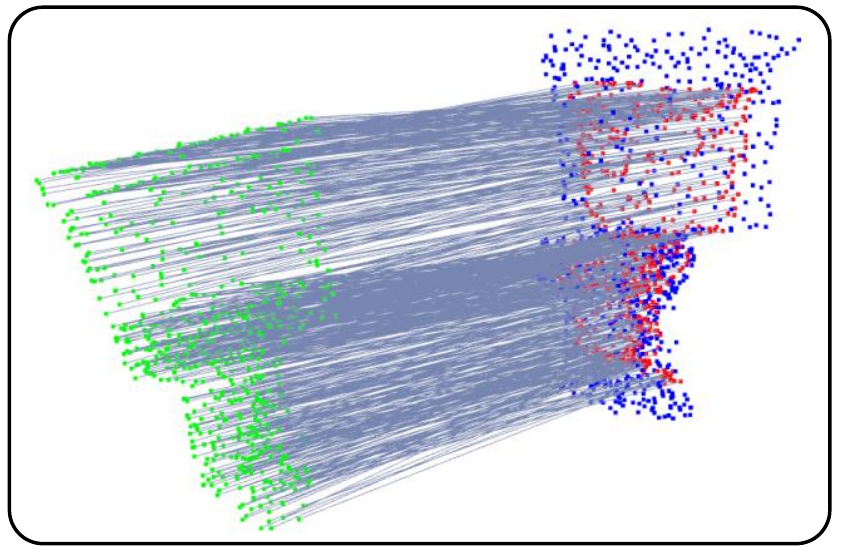}}
    		& \tabincell{c}{\includegraphics[width=1.0in]{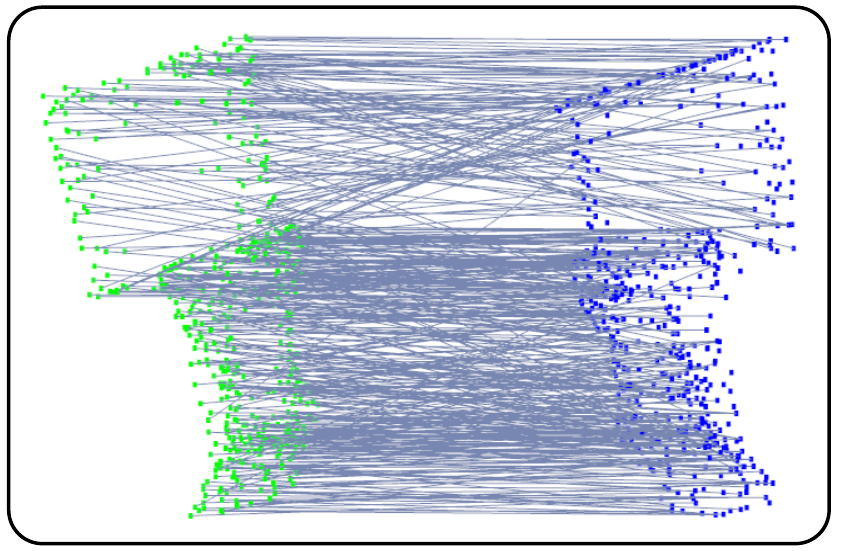}} 
    		& \tabincell{c}{\includegraphics[width=1.0in]{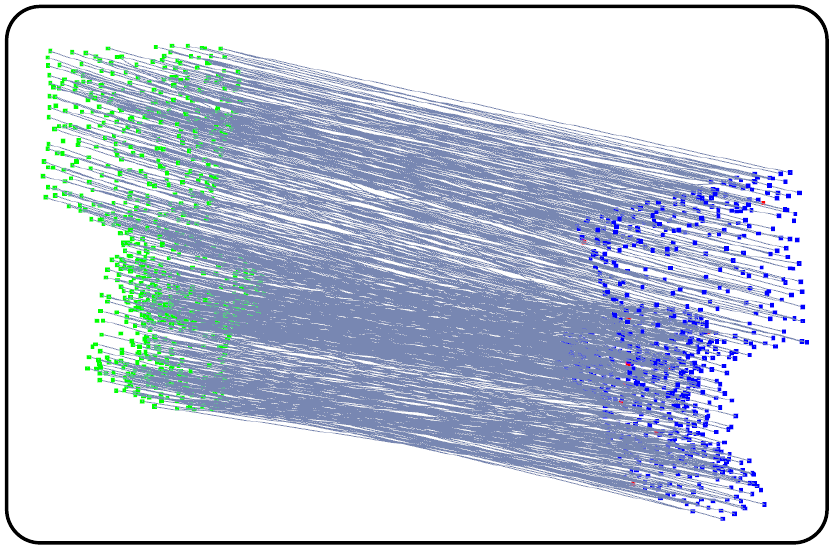}} \\
    		\bottomrule
    	\end{tabular}
	}
	\caption{Comparison of learning-based point cloud registration methods based on ``soft'' matching and ``hard'' matching. Points with different colors indicate the source (green), target (blue), and virtual points (red). Connection line indicates the correspondences. $m_{ij}$ is the entry of matching matrix $\mathbf{M}\in \mathbb{R}^{N_\mathcal{X} \times N_\mathcal{Y}}$, where $N_\mathcal{X}$, $N_\mathcal{Y}$ are size of the source and target. (The matching result of ours is returned by $\text{SHM}_\text{DCP}$.)}
	\label{tab:introduction}
\end{table*}

Moreover, PPM is defined in the discrete space, and the hard assignment algorithm is non-differentiable, which is fatal for the deep learning pipeline.
To address this issue, soft matching-based methods are proposed.
They relax the discrete matching matrix into a continuous one, where each entry is either zero or one, and then the virtual points are achieved by performing weighted average to replace the real corresponding points.
However, since the geometric constraint is ignored, the network does not learn the underlying physics, which results in an inherent ambiguity making the distribution of the virtual points degenerate seriously. 
DCP \cite{wang_dcp_iccv_2019} is a typical soft matching method, which suffers from this drawback as shown in \tabref{tab:introduction}.
RPMNet \cite{yew_rpmnet_cvpr_2020} applies the Augmented-Sinkhorn algorithm replying to outliers in the soft matching process, where the ``trash bin'' strategy \cite{superglue_paul_cvpr_2020} is employed. However, the degeneration has not been remitted.
To avoid this ambiguity trap, DeepVCP \cite{lu_deepvcp_iccv_2019} takes an unconvincing assumption, \ie accurate initial motion parameters are provided as prior. 
PRNet \cite{wang_prnet_nips_2019} and CorrNet3D \cite{zeng_corrnet3d_cvpr_2021} advocate turning the learned soft matching matrix to a hard matrix by taking the most similar points as the corresponding points of the source points. However, this strategy leads to the drawback of one-to-many matching.

In this paper, we first theoretically analyze the inherent ambiguity in these soft matching-based methods, and then devote to achieving the real PPM matrix to handle the outliers and prevent ambiguity synchronously for robust 3D point cloud registration.
However, as mentioned above, two problems block our pace. 1) Although it is well-known that the full rank permutation matrix can be solved by existing hard assignment algorithms, how to solve a PPM has yet not been explored; 2) PPM is defined in the discrete space, which is non-differentiable. 
To resolve these issues, we design a dedicated soft-to-hard (S2H) matching procedure consists of \texttt{S-step:} solving the soft matching matrix, and \texttt{H-step:} projecting this soft matrix to the PPM.
Specifically, we propose to augment the profit matrix before the hard assignment to solve an augmented permutation matrix, which is cropped to achieve the final real PPM. 
Moreover, to guarantee end-to-end learning, we supervise the learned final PPM but propagate the gradient to the soft matrix instead.
Our matching procedure can be easily integrated with existing 3D registration frameworks, which has been verified in DCP, RPMNet, and DGR \cite{choy_dgr_cvpr_2020}. 
Extensive experiments on benchmark datasets show that our method achieves state-of-the-art performance.

Our main contributions can be summarized as follows:
\begin{itemize}
\setlength{\itemsep}{0pt}
\setlength{\parsep}{0pt}
\setlength{\parskip}{0pt}
    \item We theoretically analyze the inherent ambiguity in the soft matching-based methods, which causes serious degeneration of the learned virtual corresponding points.
	\item We propose a novel S2H matching procedure to learn the PPM, which handles the outliers and prevents ambiguity synchronously. This matching procedure not only solves a real PPM, but also guarantees end-to-end learning.
	\item Remarkable performance on benchmark datasets verifies the proposed method, which achieves state-of-the-art performance in robust 3D point cloud registration.
\end{itemize}

\section{Related Works}
\label{sec:related_works}
Herein, we briefly review the learning-based point cloud registration methods. More detailed summaries have been provided in \cite{ruslinkiewicz_efficientVariantsIcp_3DDIM_2001,Fran_prc_2015,zhang_review_vrih_2020}.

\noindent\textbf{Correspondences-free methods:} 
These methods estimate the rigid motion by comparing the global representations of input two point clouds, and generally consist of two stages: global feature extraction and rigid motion estimation. 
PointNetLK \cite{aoki_ptlk_cvpr_2019} utilizes PointNet \cite{charles_pointnet_cvpr_2017} to extract global features, and then a modified LK algorithm is applied to solve the rigid motion. 
From the perspective of reconstruction, Huang \etal \cite{huang_featuremetric_cvpr_2020} utilize an encoder-decoder structure network to learn a more comprehensive global feature.

\noindent\textbf{Correspondences-based methods:} 
These methods estimate the rigid motion based on correspondences, which generally consists of feature extractor, correspondences building, and rigid motion estimation modules. 
For \textit{feature extractors}, various well-designed networks are used, such as set abstraction module \cite{charles_pointnet2_nips_2017,yew_3dfeatnet_eccv_2018}, DGCNN \cite{wang_dgcnn_tog_2019,wang_dcp_iccv_2019}, FCGF \cite{choy_fcgf_iccv_2019,choy_dgr_cvpr_2020},  globally informed 3D local feature \cite{PPFNet_Deng_2018_CVPR}, KPConv \cite{thomas_kpconv_iccv_19,D3Feat_Bai_2020_CVPR}, capsule network \cite{zhao_qec_eccv_20} and various rotation invariant features \cite{PPFFolding_Deng_2018_ECCV,perfect_Gojcic_cvpr19}. 
With the rise of deep learning, learning-based feature extractors have approached standard components, which can be integrated easily.
For \textit{correspondence building}, both soft matching-based \cite{lu_deepvcp_iccv_2019,wang_dcp_iccv_2019} and hard matching-based \cite{wang_prnet_nips_2019} methods are representative. \cite{yew_rpmnet_cvpr_2020,predator_Huang_2021_CVPR} build correspondences on the identified inliers only. 
Besides, reliable correspondences selection is the widely adopted subsequent step. It is achieved by learning the reliability weight of each initial correspondence \cite{pais_3dregnet_cvpr_2020,choy_dgr_cvpr_2020,probst_consensusMax_cvpr_2019} or selecting consistent correspondences \cite{PPFNet_Deng_2018_CVPR,PointDSC_Bai_2021_CVPR}.
For \textit{motion estimation}, Procrustes \cite{gower_procrustes_1975} is the most widely used algorithm in  \cite{wang_prnet_nips_2019,yew_rpmnet_cvpr_2020}. Recently, regressing the motion parameters directly has become a new hot spot \cite{pais_3dregnet_cvpr_2020}.

\section{Preliminaries} \label{sec:Pro_definition}
Given the source point cloud $\mathcal{X} = [\boldsymbol{x}_i]_{3 \times N_\mathcal{X}}$ and the target point cloud $\mathcal{Y}=[\boldsymbol{y}_j]_{3 \times N_\mathcal{Y}}$, where $N_\mathcal{X}$ and $N_\mathcal{Y}$ represent the sizes of the two point clouds, 3D point cloud registration aims at solving a rigid motion to best align $\mathcal{X}$ with $\mathcal{Y}$. Here, we model the rigid motion by the rotation matrix $\mathbf{R} \in SO(3)$ and the translation vector $\mathbf{t} \in \mathbb{R}^{3}$. 
Since the Procrustes algorithm \cite{gower_procrustes_1975} can optimally solve the rigid motion based on the correspondences, point matching becomes crucial, which is formulated as searching for a matching matrix $\mathbf{M}=[m_{ij}]_{N_\mathcal{X} \times N_\mathcal{Y}}$ to relate the source and target point clouds, and the entry $m_{ij}=1$ or $0$ indicates point $\boldsymbol{x}_i$ and point $\boldsymbol{y}_j$ are correspondence or not.

Ideally, $\mathcal{X}$ and $\mathcal{Y}$ are consistent, \ie points in $\mathcal{X}$ and $\mathcal{Y}$ are exactly one-to-one correspondence. 
In this case, the matching matrix $\mathbf{M}$ is a permutation matrix, \ie $m_{ij}\!\in\! \{0,1\}$, $\sum_{i=1}^N m_{ij}\!=\!1, \forall j $ and $\sum_{j=1}^N m_{ij}\!=\!1, \forall i$, where $N_\mathcal{X} \!=\!N_\mathcal{Y}\!=\!N$.
However, in practical applications, outliers always exist without corresponding points. They challenge the matching problem, and turn it into a special assignment problem while the desired matching matrix becomes a PPM
for outliers pruning. We reformulate this special PPM $\mathbf{M}$ as,
\begin{equation}
\begin{aligned} 
\{ \mathbf{M} & \ |\  m_{ij} \in \{0,1\}; \\ 
    & \sum\nolimits_{i=1}^{N_\mathcal{X}} m_{ij} \!\in\! \{0,1\}, \forall j; \sum\nolimits_{j=1}^{N_\mathcal{Y}} m_{ij} \!\in\! \{0,1\}, \forall i \}.
\label{Eq:reformulation}
\end{aligned}
\end{equation}

If the point is an inlier, the sum of the corresponding row/column equals to 1. Otherwise, the point is an outlier and the sum of the corresponding row/column equals to 0. 

\subsection{Ambiguity in soft matching-based methods}
As introduced above, different from the hard matching-based methods that build correspondences on the real points, the soft matching-based methods use the virtual corresponding points instead.
Specifically, these soft matching-based methods generate a ``soft map'' between the source and target, \ie $\boldsymbol{x}_i \in \mathcal{X}$ is assigned to $\mathcal{Y}$ by a probability vector. In this case, the matching matrix $\mathbf{M}$ becomes a soft probability matrix $\mathbf{P}$.
In DCP \cite{wang_dcp_iccv_2019}, $\mathbf{P}$ is a single stochastic matrix, where $p_{ij} \in [0,1]$, $\sum_{i=1}^{N_{\mathcal{X}}} p_{ij} = 1, \forall j $. 
In RPMNet \cite{yew_rpmnet_cvpr_2020}, $\mathbf{P}$ is optimized to a \underline{p}artial \underline{d}oubly \underline{s}tochastic \underline{m}atrix (notated as PDSM) by the Augmented-Sinkhorn algorithm, where $p_{ij} \in [0,1]$, $\sum_{i=1}^{N_{\mathcal{X}}} p_{ij} \leq 1, \forall j $ and $\sum_{j=1}^{N_{\mathcal{Y}}} p_{ij} \leq 1, \forall i$.
Then, the virtual corresponding points $\mathcal{Y}^{\prime}$ are obtained by performing weighted average on $\mathcal{Y}$ using $\mathbf{P}$, \ie $\mathcal{Y}^{\prime}=\mathcal{Y}\mathbf{P}^{\mathrm{T}}$.

However, there is an ambiguity trap here. The geometric constraints and underlying physics are ignored by relaxing the hard matching matrix to a soft probability matrix, hence, the distribution of virtual corresponding points is not unique resulting in serious degeneration. 
We can conclude as follows. The theoretical proof and more analysis are provided in \textit{supplementary materials}.

\begin{theorem}
	Considering the consistent subset point clouds $\mathbf{X}$, $\mathbf{Y}$ with ground truth motion $\mathbf{R}$, $\mathbf{t}$, there exists more than one soft matching matrix $\mathbf{P}$ satisfying $\mathbf{Y}\mathbf{P}^{\mathrm{T}} =\mathbf{R}\mathbf{X}+\mathbf{t}$.
	\label{theorem: theorem1}
\end{theorem}

Since there will be an infinite number of virtual point cloud distributions corresponding to the same rigid motion, this inherent ambiguity will cause serious degeneration of virtual points as shown in \tabref{tab:introduction}. 
Essentially, the process of weighted average can also be regarded as a special deformation of the point cloud. 
Although a seemingly good transformation estimation is obtained \cite{yew_rpmnet_cvpr_2020}, this process violates the rigid motion assumption, which cannot be supported by the Procrustes algorithm \cite{gower_procrustes_1975}.

\section{Soft-to-Hard Matching for Registration} 
\label{sec:method}

\begin{figure*}[!t]
	\centerline{\includegraphics[width=0.8\linewidth]{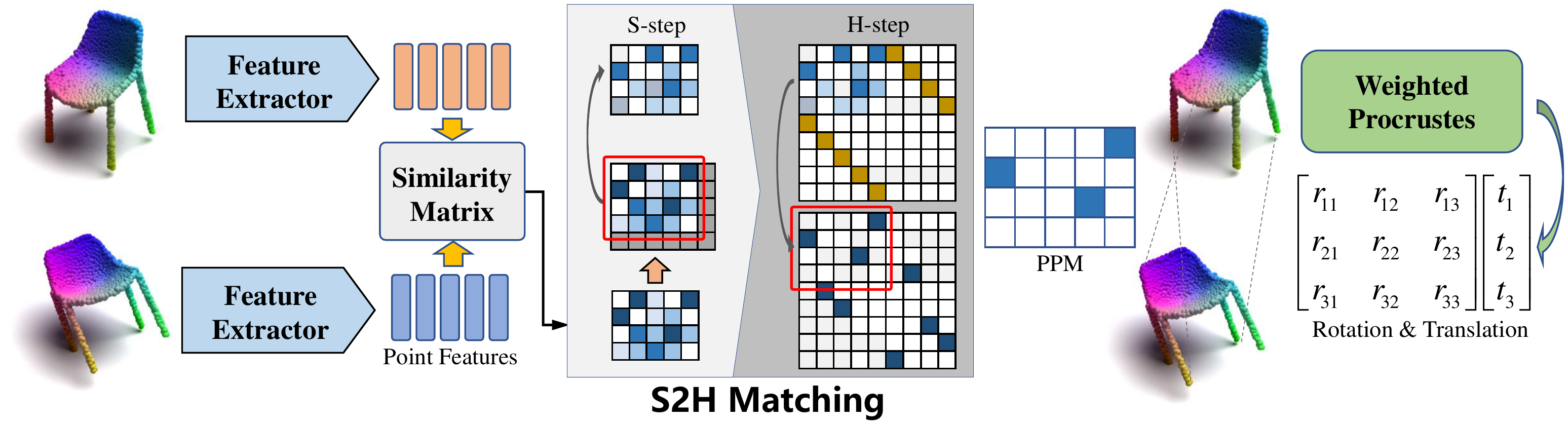}}
	\caption{Illustration of the S2H matching procedure in registration pipeline. 
	Given the source and target point clouds, the similarity matrix is obtained based on the point features. Then, S2H is applied for the final PPM. Finally, the rigid motion is estimated by the weighted Procrustes.
	}
	\label{Fig:network}
\end{figure*}

In this paper, we devote to solving a PPM to handle the outliers and prevent the ambiguity synchronously and thus design a meticulous S2H matching procedure. In this section, to clearly present S2H and how to integrate it into existing registration pipeline, we give a pipeline example as shown in \figref{Fig:network}, which consists of three core modules: feature extractor and similarity matrix solving, S2H matching, and rigid transformation estimation. Finally, the proposed loss function is also presented herein. 

\subsubsection{Feature Extractor and Similarity Matrix Solving.}

To achieve robust point matching, distinguished descriptors are crucial. With the rise of deep learning technique, many standard modules for deep features are proposed. These feature extractors can be easily integrated into our robust point cloud registration pipelines according to different requirements.

We denote the point features of the source and target as $\Phi_\mathcal{X} \in \mathbb{R}^{N_\mathcal{X}\times c}, \Phi_\mathcal{Y} \in \mathbb{R}^{N_\mathcal{Y}\times c}$. $c$ is the feature dimension. Then, based on a certain similarity metric \eg scale dot product attention \cite{vaswani_attention_nips_2017}, the similarity matrix is returned as 
$\mathbf{S}=[s_{ij}]_{N_\mathcal{X}\times N_\mathcal{Y}}$, where each entry $s_{ij}$ represents the similarity between points $\boldsymbol{x}_i \in \mathcal{X}$ and $\boldsymbol{y}_i \in \mathcal{Y}$.

\subsubsection{S2H Matching.}
A well-known solution to hard matching is to formulate it as a special assignment problem, \ie zero-one integer programming problem.
However, two natural but challenging problems exist for learning-based pipeline. 1) Existing integer programming algorithms usually achieve a row or column full rank permutation matrix ($\text{rank}(\mathbf{M})\!=\!\min(N_\mathcal{X},N_\mathcal{Y})$) rather than a PPM. It means all points of the source or target will be assigned corresponding points without distinguishing inliers and outliers. 
2) PPM is defined in the discrete space, which is non-differentiable. This characteristic is fatal for the deep learning pipeline. 
To solve these problems, we propose to augment the profit matrix to solve the PPM and design an S2H matching procedure for end-to-end learning as follows.

$\bullet$ \textbf{Augmenting the profit matrix to solve PPM:}
Conventionally, the matching task is formulated as a zero-one integer programming problem taking the similarity matrix $\mathbf{S}$ as the profit matrix, \ie,
\begin{equation}
\mathbf{M}^{*} = \arg \mathop {\max }\limits_{\mathbf{M} \in \mathcal{M}_{N}}  {<\mathbf{M},\mathbf{S}>}_{F},
\label{formulation1}
\end{equation} 
where $\mathcal{M}_{N}$ denotes the set of partial permutation matrices. ${<\mathbf{M},\mathbf{S}>}_{F}=\mathrm{trace}({\mathbf{M}}^{\mathbf{T}} \mathbf{S})$ denotes the (Frobenius) inner product.
Note that the traditional assignment algorithm will achieve a full rank permutation solution. 
In response to outliers, we propose to augment the profit matrix by adding additional rows and columns to solve an augmented permutation matrix. 
Then the PPM will be returned by cropping this augmented permutation matrix as shown in \figref{Fig:network}. Thus, the following two questions should be addressed properly.

\textit{1. How many rows and columns should be added?}
In the Augmented-Sinkhorn algorithm \cite{yew_rpmnet_cvpr_2020}, which solves a soft matching matrix, one row and one column are added as ``trash bin''. 
However, for partial permutation matching, the number of rows and columns augmented to the profit matrix is more crucial since it implies the upper bound of the number of outliers potentially.
Thus, we propose to supplement the original ${N_\mathcal{X} \times N_\mathcal{Y}}$ profit matrix to a $(N_\mathcal{X}+N_\mathcal{Y}) \times (N_\mathcal{X}+N_\mathcal{Y})$ matrix, which is a maximum redundant operation replying to the case of all points are outliers. Specifically, as shown in \figref{Fig:network}, left upper is the original matrix, right upper block is a ${N_\mathcal{X} \times N_\mathcal{X}}$ diagonal matrix and left bottom block is a ${N_\mathcal{Y} \times N_\mathcal{Y}}$ diagonal matrix, and right bottom block is a ${N_\mathcal{Y} \times N_\mathcal{X}}$ zero matrix.

\textit{2. What value to set?} 
As aforesaid, the right upper block and the left bottom block are two diagonal matrices, what values should be set to these diagonal positions? Note that these values are roughly used as thresholds to distinguish outliers and inliers potentially. 
Meanwhile, we observe that in the profit matrix, if the row/column corresponds to an outlier, the entries in this row/column approximate a uniform distribution with low value, which means the outliers have no similar points in another point cloud.
Otherwise, if the row/column corresponds to the inlier, in this row/column, the entries are close to a unimodal distribution, which means the inlier has only one similar point in another point cloud ideally.
Thus, we propose to self-adaptively fill the diagonal position with $\sigma$ according to the corresponding row/column of the input profit matrix: $\sigma \!=\! {1}/\operatorname{var}(\boldsymbol{v})$,
$\boldsymbol{v}$ is the corresponding row/column vector, and $\text{var}(\cdot)$ is the variance function.
In this case, for outlier, the filled value is large, which enforces the learned augmented permutation matrix to make the value of this position as one to gain more profit. 
For inlier, the filled value is low, the value of this position in the learned augmented permutation matrix is enforced to zero.

$\bullet$ \textbf{S2H matching in end-to-end learning:}
Since $\mathbf{M}$ is defined in the discrete space, and the integer programming algorithm is non-differentiable, we design the S2H matching procedure to guarantee end-to-end learning. 
Specifically, \texttt{S-step} learns a soft matrix and \texttt{H-step} projects this soft matrix to discrete solution space for final PPM. %

\texttt{S-step:} 
To deal with the outliers, we use Augmented-Sinkhorn \cite{yew_rpmnet_cvpr_2020} to obtain a soft matrix by adding an additional row and column of ones to the input matrix during the Sinkhorn normalization. 
In Augmented-Sinkhorn, the additional row and column are regarded as ``trash bin'', and the matching weights of outliers are expected to ``flow'' to these additional row and column to distinguish outliers and inliers.
Specifically, given the obtained similarity matrix $\textbf{S}\in \mathbb{R}^{N_\mathcal{X} \times N_\mathcal{Y}}$, this soft matrix $\textbf{P}\in \mathbb{R}^{N_\mathcal{X} \times N_\mathcal{Y}}$ is achieved by cropping the output $(N_\mathcal{X}+1) \times (N_\mathcal{Y}+1)$ matrix as shown in \figref{Fig:network}. $\textbf{P}$ is a PDSM.

\texttt{H-step:} The resultant PDSM $\mathbf{P}$ is still a soft matrix, and we project it to PPM by applying the proposed profit matrix augmenting strategy to $\mathbf{P}$ and solving this zero-one integer programming problem. 
In our implementation, we chose the classical Hungarian algorithm \cite{kuhn1955hungarian} to solve this assignment problem. 
After cropping the output augmented matrix, the final PPM $\mathbf{M}\in \mathbb{R}^{N_\mathcal{X} \times N_\mathcal{Y}}$ is obtained. 

\begin{figure}[!t]
	\centerline{\includegraphics[width=1.0\linewidth]{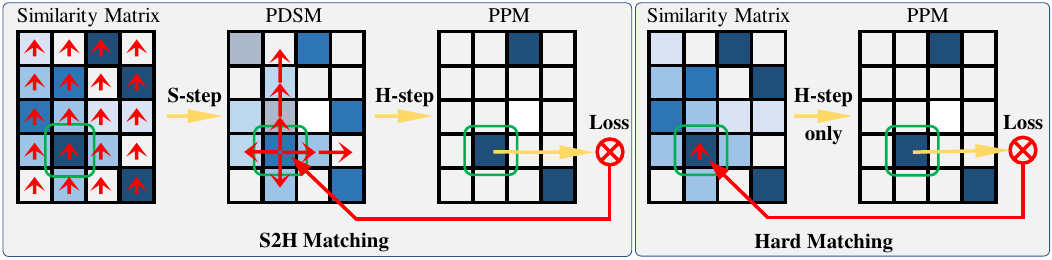}}
	\caption{Comparison of our S2H matching (Left) and hard matching (Right). Yellow and red lines represent the forward and backward propagation respectively. 
	}
	\label{Fig:Hungarian_only}
\end{figure}

For a clearer understanding, we stress the ingeniousness of the proposed S2H matching structure from two folds:

\textit{1. End-to-end learning.}
We propose a deceptive operation as shown in the left of \figref{Fig:Hungarian_only} to guarantee end-to-end learning. 
During the forward propagation, the loss is calculated based on the learned PPM $\mathbf{M}$. 
However, during the backward propagation, the gradient is not propagated to the PPM, but directly skipped to the learned PDSM $\mathbf{P}$ instead.
This ingenious structure guarantees the accuracy of the calculated loss and the backward propagation simultaneously.  

\textit{2. Hard Matching vs. S2H Matching.} 
To solve a PPM, we give a more straightforward hard matching method in the right of \figref{Fig:Hungarian_only}, which can also make sense by learning $\mathbf{M}$ considering the augmented similarity matrix as the profit matrix.
Nonetheless, there is an obvious local supervision risk in this case. That is, the gradient will be propagated to the input similarity matrix directly if only use the hard matching.
And the correlation of entries, which is considered in the integer programming process, is ignored in backward propagation.
In other words, the loss calculated from $m_{ij}$ can only supervise $s_{ij}$. This will result in only a few sparse points being supervised by the ground truth and the remaining positions will be trained without supervision.
For example, assuming $m_{ij}=1$, the gradient will be propagated to $s_{ij}$ directly, and enforce the feature extractor to enlarge $s_{ij}$.
However, the positions where $m_{ij}=0$ are not supervised (refer to \textit{loss function section} for more details).
Hence, the corresponding point features will not suppress their similarity. 
An extreme result is that all entries of the similarity matrix are very large since the point features lose the distinctiveness and each point in $\mathcal{X}$ is very similar to all points in $\mathcal{Y}$. 
This will result in a divergence of training.

Our S2H solution effectively avoids the local supervision risk by propagating the gradient to all entries in \texttt{S-step}, \ie the correlation of all entries is reconsidered.
As shown in the left of \figref{Fig:Hungarian_only}, during the backward propagation, the similarity of the correct correspondence will be boosted, meanwhile the similarity of the incorrect correspondence will be effectively suppressed.

\subsubsection{Weighted Procrustes.}
When we get a PPM $\mathbf{M}$, the corresponding point set of $\mathcal{X}$ could be achieved as $\mathcal{Y}^{\prime}= \mathcal{Y}\mathbf{M}^\mathrm{T}$. 
However, the corresponding points of outliers in $\mathcal{Y}^{\prime}$ are obtained as $[0,0,0]^\mathrm{T}$, which should be filtered out when estimating the transformation. 
To this end, the weighted Procrustes algorithm is used here. 
Given $\mathcal{X}$, $\mathcal{Y}^{\prime}$, the weight vector $\boldsymbol{w}$ is obtained as $\boldsymbol{w}=\sum_{j=1}^{N_\mathcal{Y}}{m_{ij}}$, where $\boldsymbol{w} \in \{0,1\}^{N_\mathcal{X}}$.

Then, $\boldsymbol{w}$ is normalized to $\boldsymbol{\bar{w}}$. Inspired by DGR \cite{choy_dgr_cvpr_2020}, the rigid motion is computed as follows: $\mathbf{H} = \mathcal{Y}^{\prime}\mathbf{KWK}\mathcal{X}^\mathrm{T}$,
where $\mathbf{W}=\mathrm{diag}(\boldsymbol{\bar{w}})$, $\mathbf{K}=\mathbf{I}-\sqrt{\boldsymbol{\bar{w}}}{\sqrt{\boldsymbol{\bar{w}}}}^\mathrm{T}$. $\mathbf{I}$ is an identity matrix. 
Then, $\mathbf{R} = \mathbf{U} \mathbf{E} \mathbf{V}^\mathrm{T}$, $\mathbf{t} = (\mathcal{Y}^{\prime}-\mathbf{R}\mathcal{X})\mathbf{W}\mathbf{1}$,
where $\mathbf{U}\mathbf{D}\mathbf{V}^\mathrm{T} = \text{SVD}(\mathbf{H})$, $\mathbf{1}=(1,...,1)^\mathrm{T}$ and $\mathbf{E}=\mathrm{diag}(1,...,1,\det{(\mathbf{U})}\det{(\mathbf{V})})$. 

\subsubsection{Loss function.} \label{sec:loss}
In this paper, we supervise the matching matrix directly, which is defined as:
\begin{equation}
\mathcal{L}_1 =  - \frac{{\sum_{i=1}^{N_\mathcal{X}} \sum_{j=1}^{N_\mathcal{Y}} \left( {{m}_{ij}^\text{pred} {m}_{ij}^\text{gt}} \right)}}{{\sum_{i=1}^{N_\mathcal{X}} \sum_{j=1}^{N_\mathcal{Y}} \left( {{m}_{ij}^\text{gt}} \right)}},
\label{Eq:loss1}
\end{equation}
where the superscript ``$\text{pred}$'' and ``$\text{gt}$'' represent the prediction and ground truth respectively. 
It is noteworthy that when $m_{ij}^\text{gt}=1$, $m_{ij}^\text{pred}$ is enforced to 1. But when $m_{ij}^\text{gt}=0$, $m_{ij}^\text{pred}$ diverges without supervision. 
Moreover, due to the introduction of augmentation operation, all points tend to be labeled as outliers, which makes the learned PPM close to a full zero matrix.
To avoid this degeneracy, we encourage the number of inliers by:

\begin{equation}
\mathcal{L}_2 =  -\frac{\sum_{i=1}^{N_\mathcal{X}} \sum_{j=1}^{N_\mathcal{Y}}(m_{ij}^{\text{pred}})}{N_\mathcal{X}+N_\mathcal{Y}}.
\end{equation}

Besides, we also supervise the final rigid motion, \ie,
\begin{equation}
\mathcal{L}_3 = \|{\mathbf{R}^{\text{gt}}}^{\mathrm{T}}\mathbf{R}^{\text{pred}}-\mathbf{I}_3\|_2 + \|\mathbf{t}^{\text{gt}}-\mathbf{t}^{\text{pred}}\|_2,
\end{equation}
where $\mathbf{I}_3$ is a $3\times3$ identity matrix.
Then, our final loss function is reached as $\mathcal{L} = \lambda_1 \mathcal{L}_1 + \lambda_2 \mathcal{L}_2 + \lambda_3 \mathcal{L}_3$,
the trade-off parameters are set to $\lambda_1 = \lambda_2 = \lambda_3 = 1$ in this paper.

\section{Experiments and Evaluation} \label{sec:experiment}

In this section, we evaluate the proposed S2H matching procedure in several representative point cloud registration frameworks including DCP, RPMNet and DGR, on benchmark datasets including ModelNet40 \cite{wu_modelnet40_cvpr_2015}, 3DMatch \cite{zeng_3dmatch_cvpr_17} and KITTI \cite{geiger_kitti_rr_13}.

\noindent\textbf{Implementation details.} 
To validate the proposed matching method can be generally integrated, we evaluate S2H matching within three typical frameworks, \ie DCP, RPMNet, and DGR, notated as $\text{SHM}_\text{DCP}$, $\text{SHM}_\text{RPMNet}$, and $\text{SHM}_\text{DGR}$ respectively. These three frameworks are typical and representative, where DCP and RPMNet are soft matching-based methods using different feature extractors, and mainly concentrate on the synthetic dataset, ModelNet40. DGR is a hard matching-based method focusing on large-scale real datasets, 3DMatch and KITTI.  
Note that PRNet \cite{wang_prnet_nips_2019} also inherits the framework of DCP, which is compared with $\text{SHM}_\text{DCP}$ herein.
The complete loss is used in $\text{SHM}_\text{DCP}$, $\text{SHM}_\text{RPMNet}$. 
Only $\mathcal{L}_3$ is used in $\text{SHM}_\text{DGR}$. 
Refer to \textit{supplementary materials} for more details.

\subsection{Evaluation on synthetic dataset: ModelNet40} \label{sec:exp:modelnet40}

In this section, we validate our proposed S2H matching procedure with $\text{SHM}_\text{DCP}$, $\text{SHM}_\text{RPMNet}$ on a synthetic dataset, ModelNet40. 
Following DCP and RPNet, we construct a point cloud by randomly sampling 1024 points, and then apply a rigid transformation to this point cloud, where the rotation and translation are uniformly sampled from $[0^ \circ, 45^ \circ]$ and $[-0.5, 0.5]^3$ respectively along each axis. 
Next, we randomly sample 768 points from the original point cloud and the transformed point cloud as the source and target to ensure the random distribution of the outliers.

In addition, following DCP and RPMNet, we test ours on two different dataset settings. 1) {Unseen categories} (\textit{clean}): ModelNet40 will be divided into training and test splits based on the object category, \ie the first 20 categories are selected for training and the rest categories for testing. 2) {Noisy data} (\textit{noisy}): For robustness testing, random Gaussian noise (\ie, $\mathcal{N}(0,0.01)$) is added to each point, while the sampled noise out of the range of $[-0.05,0.05]$ will be clipped. The dataset splitting strategy is the same as \textit{clean}.

\noindent\textbf{Matching.} Accurate correspondences estimation is crucial for robust point cloud registration. Here, we evaluate the constructed correspondences for a clear comparison. 

$\bullet$ \textbf{Metric:} We calculate the discrepancy between the predicted and the ground truth corresponding points. The predicted corresponding points $\mathcal{Y}_\text{pred}^{\prime}$ of $\mathcal{X}$ can be obtained by two methods. First, based on the \textit{{predicted matching matrix}} $\mathbf{M}^\text{pred}$, $\mathcal{Y}_\text{pred}^{\prime}$ can be obtained by $\mathcal{Y}_\text{pred}^{\prime}={\mathcal{Y}\mathbf{M}^{\text{pred}\mathrm{T}}}$. 
Second, inspired by the iteration strategy in ICP, $\mathcal{Y}_\text{pred}^{\prime}$ can be obtained based on the \textit{{predicted transformation}} $\{\mathbf{R}^\text{pred},\mathbf{t}^\text{pred}\}$ and \textit{{nearest neighbor principle}}, \ie $\mathcal{Y}_\text{pred}^{\prime}=\text{NN}_{\mathcal{Y}}(\mathbf{R}^\text{pred} \mathcal{X} + \mathbf{t}^\text{pred})$, where $\text{NN}_\mathcal{Y}(\cdot)$ solves the nearest neighbor point in $\mathcal{Y}$. 
Given the ground truth corresponding points $\mathcal{Y}_\text{gt}^{\prime}$, the root mean squared error (RMSE) and mean absolute error (MAE) in Euclidean distance between $\mathcal{Y}_\text{pred}^{\prime}$ and $\mathcal{Y}_\text{gt}^{\prime}$, notated as RMSE(dis) and MAE(dis) are presented.

We also report the matching recall $(\%)$ based on a proposed self-adaptive threshold, ${\tau_i} = ({1}/{K}) {\sum_{j = 1}^K {{d_{\boldsymbol{y}_{\pi(i)}^{\prime}, \boldsymbol{y}_j}}}}$, where $\boldsymbol{y}_{\pi(i)}^{\prime}$ is the correct corresponding point of $\boldsymbol{x}_{i}$,
$\boldsymbol{y}_{j} \in \text{KNN}_\mathcal{Y} (\boldsymbol{y}_{\pi(i)}^{\prime})$, 
$\text{KNN}_\mathcal{Y}(\cdot)$ solves the K-nearest neighbor points (exclude the self-point) in $\mathcal{Y}$ with pre-defined parameter $K$,
${d_{\boldsymbol{y}_{\pi(i)}^{\prime},\boldsymbol{y}_{j}}}$ is the distance between $\boldsymbol{y}_{\pi(i)}^{\prime}$ and $\boldsymbol{y}_{j}$.  
To sum up, for the $i$-th point, $\tau_i$ is computed as the average distance of K-nearest points around the correct corresponding point in $\mathcal{Y}$. If the distance between the correct corresponding point and the predicted one is less than the threshold, this pair will be confirmed as a correct matching pair.

\begin{table}[!h]
	\renewcommand\arraystretch{1.0}
	\begin{center}
	    \resizebox{0.99\linewidth}{!}{
			\begin{tabular}{lcccc|cccc}
				\toprule
				\multirow{2}*{{\textbf{Methods}}}&\multicolumn{2}{c}{$\mathbf{RMSE(dis)}$ $\downarrow$}&\multicolumn{2}{c|}{$\mathbf{MAE(dis)}$ $\downarrow$}&\multicolumn{2}{c}{$\mathbf{RMSE(dis)}$ $\downarrow$}&\multicolumn{2}{c}{$\mathbf{MAE(dis)}$ $\downarrow$} \\
				\cmidrule(r){2-3} \cmidrule(r){4-5} \cmidrule(r){6-7} \cmidrule(r){8-9}
				~&\multicolumn{1}{c}{\textit{clean}}&\multicolumn{1}{c}{\textit{noisy}}&\multicolumn{1}{c}{\textit{clean}}&\multicolumn{1}{c|}{\textit{noisy}}&\multicolumn{1}{c}{\textit{clean}}&\multicolumn{1}{c}{\textit{noisy}}&\multicolumn{1}{c}{\textit{clean}}&\multicolumn{1}{c}{\textit{noisy}}\\
				\midrule
				{DCP-v2}  &0.500  &0.466  &0.705 &0.657 &0.088 &0.095 &0.078 &0.095 \\
				{PRNet} &0.311  &0.303  &0.357 &0.351 &0.046 &0.057 &0.023 &0.036 \\
				$\text{SHM}_\text{DCP}$ &\textbf{0.106}  &\textbf{0.188}  &\textbf{0.027} &\textbf{0.082} &\textbf{0.033} &\textbf{0.053} &\textbf{0.008} &\textbf{0.019} \\
				\midrule
				{RPMNet} &0.139 &0.137 &0.182 &0.181 &0.019 &0.028 &0.004 &0.013 \\
				$\text{SHM}_\text{RPMNet}$ &\textbf{0.057} &\textbf{0.121} &\textbf{0.009} &\textbf{0.041} &\textbf{0.007} &\textbf{0.027} &\textbf{0.001} &\textbf{0.010} \\
				\bottomrule
			\end{tabular}}
	\end{center}
	\caption{Discrepancy based on predicted matching matrix ({Left}) and predicted transformation, nearest neighbor principle ({Right}).}
	\label{tab:match_dis}
\end{table}

$\bullet$ \textbf{Evaluation:}
In \tabref{tab:match_dis}, we report RMSE(dis) and MAE(dis) results in \textit{clean} and \textit{noisy}. For the discrepancy based on the predicted matching matrix, our method improves the performance with a big margin in both DCP and RPMNet frameworks. These obtained results are reasonable because the DCP-v2, RPMNet are virtual point-based methods, where the corresponding points degenerate seriously with a large distance to correct corresponding points. PRNet presents weaker matching performance due to the one-to-many matching. For the discrepancy based on the predicted transformation and nearest neighbor principle, all methods achieve superior performance, whereas $\text{SHM}_\text{DCP}$, $\text{SHM}_\text{RPMNet}$ remain the overall best performance.

\begin{figure}[!ht]
	\centerline{\includegraphics[width=1.0\linewidth]{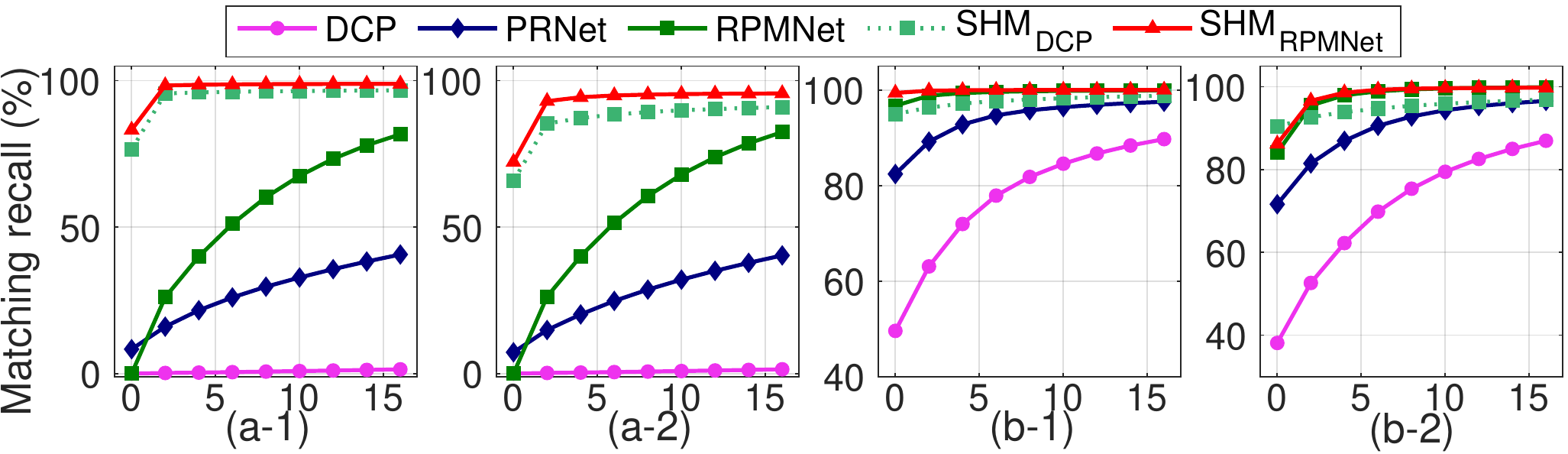}}
	\caption{Matching recall with different thresholds based on the predicted matching matrix ((a-1), (a-2)) and the predicted transformation, nearest neighbor principle ((b-1), (b-2)). (a-1), (b-1) are conducted in \textit{clean} and (a-2), (b-2) in \textit{noisy}. The X-axis is the number of K-nearest points for threshold, and $K=0$ means $\tau_i=0$.}
	\label{Fig:rec}
\end{figure}

Besides, we also draw the matching recall with different thresholds in \figref{Fig:rec}. For the results based on the predicted matching matrix, the DCP-v2, RPMNet fail to obtain accurate correspondences. And ours obtains the best results.
For the results based on the predicted transformation and nearest neighbor principle, all methods achieve better performance, and $\text{SHM}_\text{DCP}$, $\text{SHM}_\text{RPMNet}$ remain the best performance. 

\noindent\textbf{Registration.}
In this section, we evaluate the rigid motion estimation performance of $\text{SHM}_\text{DCP}$, $\text{SHM}_\text{RPMNet}$.

$\bullet$ \textbf{Metric:} Following DCP, RMSE and MAE between the ground truth and prediction in Euler angle and translation vector are used as the evaluation metrics here, notated as RMSE(R), MAE(R), RMSE(t) and MAE(t) respectively. 

\begin{figure}[!ht]
	\centerline{\includegraphics[width=1.0\linewidth]{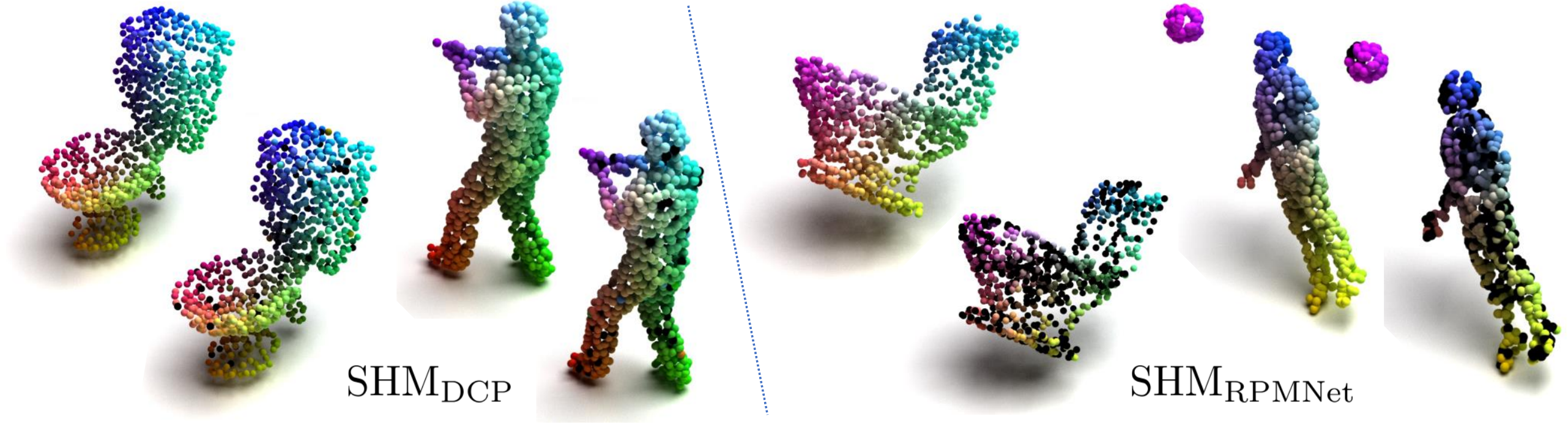}}
	\caption{For each pair of point clouds, left upper is the transformed source using the predicted rigid motion and right bottom is the target. For a more clear presentation of the outliers, we only sample the target of the input pair, \ie $N_\mathcal{X}=1024$, $N_\mathcal{Y}=768$. The same color represents the correspondences, and black indicates the abandoned points. Ours achieves advanced correspondences building and robust registration results in such challenging cases.}
	\label{Fig:Matchingresult2}
\end{figure}

$\bullet$ \textbf{Evaluation:}
The evaluation results are provided in \tabref{tab:registration}. Both
$\text{SHM}_\text{DCP}$ and $\text{SHM}_\text{RPMNet}$ outperform the hand-craft registration methods, ICP \cite{besl_icp_pami_1992}, FGR \cite{zhou_fgr_eccv_2016}. 
Furthermore, in \textit{clean}, $\text{SHM}_\text{DCP}$ achieves more accurate performance than DCP-v2 and PRNet in all metrics, $\text{SHM}_\text{RPMNet}$ is more accurate than RPMNet in all metrics. 
In \textit{noisy}, we train and test ours and all baselines with noisy data. And the methods with our S2H matching also achieves the best performance in these two frameworks in all evaluation metrics. These evaluations validate the improvement of our S2H matching, which creates a new state-of-the-art results. 
More intuitively, we provide some qualitative registration performance in \figref{Fig:Matchingresult2}.

\begin{table}[!ht]
	\renewcommand\arraystretch{1.0}
	\begin{center}
	    \resizebox{0.99\linewidth}{!}{
			\begin{tabular}{lcccccccc}
				\toprule
				\multirow{2}*{{\textbf{Methods}}}&\multicolumn{2}{c}{$\mathbf{RMSE(R)}$ $\downarrow$}&\multicolumn{2}{c}{$\mathbf{MAE(R)}$ $\downarrow$}&\multicolumn{2}{c}{$\mathbf{RMSE(t)}$ $\downarrow$}&\multicolumn{2}{c}{$\mathbf{MAE(t)}$ $\downarrow$} \\
				\cmidrule(r){2-3} \cmidrule(r){4-5} \cmidrule(r){6-7} \cmidrule(r){8-9}
				~&\multicolumn{1}{c}{\textit{clean}}&\multicolumn{1}{c}{\textit{noisy}}&\multicolumn{1}{c}{\textit{clean}}&\multicolumn{1}{c}{\textit{noisy}}&\multicolumn{1}{c}{\textit{clean}}&\multicolumn{1}{c}{\textit{noisy}}&\multicolumn{1}{c}{\textit{clean}}&\multicolumn{1}{c}{\textit{noisy}}\\
				\midrule
				\text{ICP}     &12.545 &12.723 &5.438 &5.298 &0.046 &0.045 &0.024 &0.023 \\
				\text{FGR}     &20.042 &40.829 &7.203 &21.065&0.035 &0.060 &0.019 &0.039 \\
				\midrule
				\text{DCP-v2}  &6.265  &6.347  &3.990 &4.294 &0.014 &0.016 &0.011 &0.012 \\
				\text{PRNet}   &3.532  &4.321  &1.760 &1.826 &0.013 &0.013 &0.010 &0.010 \\
				$\text{SHM}_\text{DCP}$ &\makecell[c]{\textbf{2.522}\\\scriptsize{\color{SpringGreen}+28.60\%}}  
				                  &\makecell[c]{\textbf{3.886}\\\scriptsize{\color{SpringGreen}+10.07\%}}  
				                  &\makecell[c]{\textbf{0.833}\\\scriptsize{\color{SpringGreen}+52.67\%}}   
				                  &\makecell[c]{\textbf{1.510}\\\scriptsize{\color{SpringGreen}+17.31\%}}  
				                  &\makecell[c]{\textbf{0.005}\\\scriptsize{\color{SpringGreen}+61.54\%}}  
				                  &\makecell[c]{\textbf{0.006}\\\scriptsize{\color{SpringGreen}+53.85\%}}  
				                  &\makecell[c]{\textbf{0.003}\\\scriptsize{\color{SpringGreen}+70.00\%}}   
				                  &\makecell[c]{\textbf{0.004}\\\scriptsize{\color{SpringGreen}+60.00\%}}   \\
				\midrule
				\text{RPMNet} &0.886 &1.631 &0.345 &0.565 &0.006 &0.011 &0.003 &0.004 \\
				$\text{SHM}_\text{RPMNet}$ &\makecell[c]{\textbf{0.514}\\\scriptsize{\color{SpringGreen}+41.99\%}} 
				                  &\makecell[c]{\textbf{1.456}\\\scriptsize{\color{SpringGreen}+10.73\%}} 
				                  &\makecell[c]{\textbf{0.247}\\\scriptsize{\color{SpringGreen}+28.41\%}}
				                  &\makecell[c]{\textbf{0.378}\\\scriptsize{\color{SpringGreen}+33.10\%}} 
				                  &\makecell[c]{\textbf{0.004}\\\scriptsize{\color{SpringGreen}+33.33\%}} 
				                  &\makecell[c]{\textbf{0.008}\\\scriptsize{\color{SpringGreen}+27.27\%}} 
				                  &\makecell[c]{\textbf{0.002}\\\scriptsize{\color{SpringGreen}+33.33\%}} 
				                  &\makecell[c]{\textbf{0.003}\\\scriptsize{\color{SpringGreen}+25.00\%}} \\
				\bottomrule
			\end{tabular}}
	\end{center}
	\caption{Registration perfomance on \textit{clean}, \textit{noisy}. Green notes the improvement comparing with the second-top results.}
	\label{tab:registration}
\end{table}

$\bullet$ \textbf{Time-efficiency:} 
\label{sec::time_complexity}
We counted the average inference time of learning-based methods in \tabref{Tab:times}. The experiments are conducted in \textit{noisy} on ModelNet40 using a Xeon E5-2640 v4@2.40GHz CPU and a GTX 1080Ti. 
S2H matching-based methods are slower because the integer programming algorithm is time-consuming. $\text{SHM}_\text{RPMNet}$ is slower than $\text{SHM}_\text{DCP}$ due to the iteration in RPMNet framework.

\begin{table}[!ht]
	\renewcommand\arraystretch{1.0}
	\begin{center}
		\resizebox{0.85\linewidth}{!}{
			\begin{tabular}{lccccc}
				\toprule
				\text{Methods} & \text{DCP} & \text{PRNet} & $\text{SHM}_\text{DCP}$ & \text{RPMNet} & $\text{SHM}_\text{RPMNet}$ \\
				\midrule				
				{Time(ms)} & \textbf{10.66} & 45.48 & 98.03 & 54.75 & 409.95\\
				\bottomrule
		\end{tabular}}
	\end{center}
	\caption{Inference time comparison of learning-based methods.}
	\label{Tab:times}
\end{table}

\subsection{Evaluation on real indoor data: 3DMatch} \label{sec:exp:3dmatch}
In this part, we evaluate our S2H matching in $\text{SHM}_\text{DGR}$ on real indoor dataset, 3DMatch \cite{zeng_3dmatch_cvpr_17}. $\text{SHM}_\text{DGR}$ takes the S2H matching replacing the original matching of DGR \cite{choy_dgr_cvpr_2020} (more details can be seen in \textit{supplementary materials}).
Following DGR, the input point clouds have been voxelized with the voxel size of 5cm, then each of them contains $\sim$50k points.

$\bullet$ \textbf{Metric:} 
For a fair comparison, we follow the protocols and evaluation metrics of DGR here.
The average \underline{R}otation \underline{E}rror (RE: $\text{arccos}(({trace({\mathbf{R}^\text{pred}}^{-1}\mathbf{R}^\text{gt})-1)/2})\frac{180}{\pi}$), average \underline{T}ranslation \underline{E}rror (TE: $\|\mathbf{t}^\text{pred}-\mathbf{t}^\text{gt}\|_2^2$), recall, and time-efficiency are reported. Recall indicates the ratio of successful pairwise registrations. 
Here, the successful pair is confirmed if the RE and TE are smaller than pre-defined thresholds (\ie, RE$<$$15^\circ$, TE$<$$30cm$).

\begin{table}[!ht]
	\renewcommand\arraystretch{1.0}
	\begin{center}
	\resizebox{0.99\linewidth}{!}{
		\begin{tabular}{lcccc}
			\toprule
				\textbf{Methods} & \textbf{TE}(cm) $\downarrow$ & \textbf{RE}(deg) $\downarrow$ & \textbf{Recall}(\%) $\uparrow$ & \textbf{Time}(s) $\downarrow$\\
				\midrule		
				\text{ICP}                 & 18.1    & 8.25   & 6.04   & 0.25   \\
				\text{FGR}                 & 10.6    & 4.08   & 42.7   & 0.31   \\
				\text{Go-ICP}              & 14.7    & 5.38   & 22.9   & 771.0  \\
				\text{Super4PCS}           & 14.1    & 5.25   & 21.6   & 4.55   \\
				\text{RANSAC}           & 9.16    & 2.95   & 70.7   & 2.32   \\
				\midrule
				\text{DCP-v2}              & 21.4    & 8.42   & 3.22   & \textbf{0.07}   \\
				\text{PointNetLK}          & 21.3    & 8.04   & 1.61   & 0.12   \\
				\text{DGR w/o safeguard}   & 7.73    & 2.58   & 85.2   & 0.70   \\
				\text{DGR}                 & 7.34    & 2.43   & {91.3}   & 1.21   \\
				\text{PointDSC}            & 6.55    & 2.06   & \textbf{93.28}  & 0.09   \\
				\midrule
				$\text{SHM}_\text{DGR}$    & \textbf{6.41}    & \textbf{1.75}   & 91.7   & 37.2   \\
				\bottomrule
	\end{tabular}}
	\caption{Evaluation on 3DMatch dataset.}
	\label{Tab:3dmatch}
	\end{center}
\end{table}

$\bullet$ \textbf{Evaluation:} From the \tabref{Tab:3dmatch}, we find that ICP achieves the weak registration result since the dataset contains challenging sequences with large motion while no reliable prior is provided. Super4PCS \cite{mellado_super4pcs_cgf_14}, and Go-ICP \cite{yang_goicp_pami_2015}, which are sampling-based algorithm and the variant of ICP with branch-and-bound search respectively, present similar performance here. FGR and RANSAC perform better due to the extracted point features. 3DRegNet \cite{pais_3dregnet_cvpr_2020} is also tested here, however, it does not converge, which outputs the error above 30$^{\circ}$ and 1m. DGR is the state-of-the-art learning-based method, which is designed for scene data specifically. However, DGR also takes the most similar points as the corresponding points ignoring the one-to-one matching principle. $\text{SHM}_\text{DGR}$ achieves better registration performance including transformation estimation and recall comparing with the original DGR. PointDSC \cite{PointDSC_Bai_2021_CVPR} achieves the best recall but the registration performance is weaker than $\text{SHM}_\text{DGR}$.

\subsection{Evaluation on real outdoor data: KITTI} \label{sec:exp:kitti}
We evaluate $\text{SHM}_\text{DGR}$ on the real outdoor data, KITTI \cite{geiger_kitti_rr_13}. Here, we also follow the protocols of DGR, where the evaluation metrics are identical to the 3DMatch evaluation. 
The thresholds to confirm the successful pair are set to 0.6m and 5$^\circ$. 
From \tabref{Tab:kitti}, the $\text{SHM}_\text{DGR}$ achieves the best performance with respect to rigid transformation estimation and recall, which outperforms the original DGR and the state-of-the-art method, PointDSC. 

\begin{table}[!t]
	\renewcommand\arraystretch{1.0}
	\centering
	\resizebox{0.9\linewidth}{!}{
		\begin{tabular}{lccccc}
			\toprule
			    \multirow{1}*{\textbf{Methods}}&\multicolumn{1}{c}{\textbf{TE}(cm) $\downarrow$}
               &\multicolumn{1}{c}{\textbf{RE}(deg) $\downarrow$}
               &\multicolumn{1}{c}{\textbf{Recall}(\%)$\uparrow$}
               &\multirow{1}*{\textbf{Time}(s) $\downarrow$}
				 \\

				\midrule		
				\text{FGR}    & 40.7  & 1.02   &0.2  &1.42 \\
				\text{RANSAC} & 25.9  & 1.39   &34.2 &1.37 \\
				\text{FCGF}   & 10.2  & 0.33   &98.2 &6.38 \\
				\text{DGR}    & 21.7  & 0.34   &96.9 &2.29 \\
				\text{PointDSC}&20.94 & 0.33   &98.20&\textbf{0.31} \\
				\midrule
				$\text{SHM}_\text{DGR}$  & \textbf{9.32}  & \textbf{0.28}   &\textbf{99.3} &52.4 \\
				\bottomrule
	\end{tabular}
	}
	\caption{Evaluation on KITTI dataset.}
	\label{Tab:kitti}
\end{table}

\subsection{Ablation studies} \label{sec:experiment:ablation}

\noindent\textbf{End-to-end vs. post-processing.} 
In this paper, we advocate learning the one-to-one matching in an end-to-end manner, \ie achieving the PPM in the matching stage. Oppositely, another natural idea is to solve the PPM in post-processing, \ie using only the \texttt{S-step} to learn the soft matrix during the training and adding the \texttt{H-step} during the test for the final hard matrix, PPM. 
The comparison of these two ideas is given in \tabref{tab:end2end}, where the experiments are conducted on ModelNet40. We can see that the end-to-end learning achieves better results in all metrics because more accurate loss are calculated in this pipeline.

\begin{table}[!ht]
	\renewcommand\arraystretch{1.0}
	\begin{center}
	    \resizebox{\linewidth}{!}{
			\begin{tabular}{lcccccccc}
				\toprule
				\multirow{2}*{{\textbf{Methods}}}&\multicolumn{2}{c}{$\mathbf{RMSE(R)}$ $\downarrow$}&\multicolumn{2}{c}{$\mathbf{MAE(R)}$ $\downarrow$}&\multicolumn{2}{c}{$\mathbf{RMSE(t)}(\times 10^{-3})$ $\downarrow$}&\multicolumn{2}{c}{$\mathbf{MAE(t)}(\times 10^{-3})$ $\downarrow$} \\
				\cmidrule(r){2-3} \cmidrule(r){4-5} \cmidrule(r){6-7} \cmidrule(r){8-9}
				~&\multicolumn{1}{c}{\textit{clean}}&\multicolumn{1}{c}{\textit{noisy}}&\multicolumn{1}{c}{\textit{clean}}&\multicolumn{1}{c}{\textit{noisy}}&\multicolumn{1}{c}{\textit{clean}}&\multicolumn{1}{c}{\textit{noisy}}&\multicolumn{1}{c}{\textit{clean}}&\multicolumn{1}{c}{\textit{noisy}}\\
				\midrule
				$\text{SHM}_\text{DCP}^-$  &3.325          &4.570          &1.039          &1.683          &5.056  &6.624          &3.204 &4.316 \\
				$\text{SHM}_\text{DCP}^+$  &\makecell[c]{\textbf{2.522} \\\scriptsize{\color{SpringGreen}+24.15\%}}
				                       &\makecell[c]{\textbf{3.886} \\\scriptsize{\color{SpringGreen}+14.97\%}} 
				                       &\makecell[c]{\textbf{0.833} \\\scriptsize{\color{SpringGreen}+19.83\%}} 
				                       &\makecell[c]{\textbf{1.510} \\\scriptsize{\color{SpringGreen}+10.28\%}} 
				                       &\makecell[c]{\textbf{4.780}         \\\scriptsize{\color{SpringGreen}+5.46\%}}          
				                       &\makecell[c]{\textbf{5.758} \\\scriptsize{\color{SpringGreen}+13.07\%}} 
				                       &\makecell[c]{\textbf{3.052}          \\\scriptsize{\color{SpringGreen}+4.74\%}}          
				                       &\makecell[c]{\textbf{3.774} \\\scriptsize{\color{SpringGreen}+12.56\%}} \\
				\midrule
				$\text{SHM}_\text{RPMNet}^-$  &0.760          &1.537          &0.270          &0.449          &4.927          &10.120         &2.356          &3.602 \\
				$\text{SHM}_\text{RPMNet}^-$  &\makecell[c]{\textbf{0.514} \\\scriptsize{\color{SpringGreen}+32.37\%}}
				                       &\makecell[c]{\textbf{1.456} \\\scriptsize{\color{SpringGreen}+5.27\%}} 
				                       &\makecell[c]{\textbf{0.247} \\\scriptsize{\color{SpringGreen}+8.52\%}} 
				                       &\makecell[c]{\textbf{0.378} \\\scriptsize{\color{SpringGreen}+15.81\%}} 
				                       &\makecell[c]{\textbf{3.883} \\\scriptsize{\color{SpringGreen}+21.19\%}} 
				                       &\makecell[c]{\textbf{8.421} \\\scriptsize{\color{SpringGreen}+16.79\%}} 
				                       &\makecell[c]{\textbf{2.328} \\\scriptsize{\color{SpringGreen}+1.19\%}} 
				                       &\makecell[c]{\textbf{3.145} \\\scriptsize{\color{SpringGreen}+12.69\%}} \\
				\bottomrule
			\end{tabular}}
	\end{center}
	\caption{Comparison between the end-to-end learning and post-processing setting. ``-'' indicates the post-processing, and ``+'' indicates the end-to-end learning.}
	\label{tab:end2end}
\end{table}

\noindent\textbf{Other important ablation studies.} Due to the limitation of space, we provide some other import experiments in \textit{supplementary materials}, including the robustness to different outliers generation strategy, the robustness to different outliers ratio, the influence of different loss function combinations, and more qualitative results of registration, \etc.

\section{Conclusion} \label{sec:conclusion}
In this paper, we tackle the point matching problem in robust 3D point cloud registration. First, we analyze the inherent ambiguity in soft matching-based methods. Second, to resolve the ambiguity and handle the outliers in the matching stage, we propose to learn the partial permutation matching (PPM) matrix. To address the consequent problem that PPM is defined in non-differentiable space and cannot be solved by existing hard assignment algorithms, we design a soft-to-hard matching method. 
We have validated the effectiveness by integrating it with various registration frameworks including DCP, RPMNet, and DGR and conducting extensive experiments in both synthetic data (ModelNet40) and real scan data (3DMatch and KITTI), which created a new state-of-the-art performance for robust 3D point cloud registration. In the future, we plan to extend our framework to non-rigid point cloud registration.

\section*{Acknowledgments}
This research was supported in part by National Key Research and Development Program of China (2018AAA0102803) and National Natural Science Foundation of China (61871325, 62001394, 61901387). Besides, this work was also sponsored by Innovation Foundation for Doctor Dissertation of Northwestern Polytechnical University.

\bibliography{References}
\end{document}


\maketitle

\begin{abstract}
In this supplementary material, we provide additional proof, analysis, ablation studies and qualitative results, which are important but omitted in the main paper due to space limitation. First, we present the detailed proof of the inherent ambiguity within the soft matching-based methods. Second, we provide details of the network architecture and the implementation. Third, additional experiments and ablation studies are presented including robustness to different outliers generation strategies, robustness to different outliers ratios, influence of different loss function combinations. We conclude this material with additional qualitative 3D registration results on both synthetic and real-world datasets.
\end{abstract}

\section{Inherent ambiguity in \\ the soft matching-based methods}
\label{sec:supp:ambiguity}

\begin{theorem}
	Considering the consistent subset point clouds $\mathbf{X}$, $\mathbf{Y}$ with ground truth motion $\mathbf{R}$, $\mathbf{t}$, there exists more than one soft matching matrix $\mathbf{P}$ satisfying $\mathbf{Y}\mathbf{P}^{\mathrm{T}} =\mathbf{R}\mathbf{X}+\mathbf{t}$.
	\label{theorem: theorem1}
\end{theorem}

\begin{proof}
\end{proof}
	Generally, points in $\mathbf{X} $ and $\mathbf{Y}$ are not in the same order and there exists a ground truth matching matrix $\mathbf{M}^{*}$ (\ie permutation matrix) that makes $\mathbf{X}$ and $\mathbf{Y}{\mathbf{M}^{*}}^{\mathrm{T}}$ match in sequence. Then an analytical solution of rotation matrix $\mathbf{R}^{*}$ can be solved by the Procrustes algorithm \rm\cite{gower_procrustes_1975}, we indicate these correct results by $^*$, \ie
	\begin{equation} 
	\left\{ {\begin{aligned} 
		\mathbf{H^*} &= \mathbf{X}(\mathbf{Y} {\mathbf{M^{*}}^{\mathrm{T}}})^\mathrm{T} \\
		\mathbf{H^*} &= \mathbf{U^*} \mathbf{D^*} \mathbf{V^*}^\mathrm{T} 
		\end{aligned}} \right. 
	\Longrightarrow 
	\mathbf{R^*} = \mathbf{V^*} \mathbf{U^*}^\mathrm{T},
	\label{eq:permutation_eq}
	\end{equation}
where $\mathbf{U}^{*}$, $\mathbf{D}^{*}$ and $\mathbf{V}^{*}$ are obtained by performing the singular value decomposition to $\mathbf{H}^{*}$. 
	
	
Similarly, when a certain soft matching matrix $\mathbf P$ is given, the predicted rotation matrix $\mathbf{R}$ can be solved as, 
	\begin{equation} 
	\left\{ {\begin{aligned} 
		\mathbf{H} &= \mathbf{X} (\mathbf{Y} \mathbf{P}^{\mathrm{T}})^\mathrm{T}  \\
		\mathbf{H} &= \mathbf{U} \mathbf{D} \mathbf{V}^\mathrm{T}  
		\end{aligned}} \right. 
	\Longrightarrow 
	\mathbf{R} = \mathbf{V} \mathbf{U}^\mathrm{T}.
	\label{eq:map_matrix}
	\end{equation}
	Based on the equation of $\mathbf{X} \mathbf{P} \mathbf{Y}^\mathrm{T} = \mathbf{H} =  \mathbf{U} \mathbf{D} \mathbf{V}^\mathrm{T}$, the soft matching matrix $\mathbf P$ can be solved as
	\begin{equation}
	\begin{aligned}
	\mathbf{P} = {{\mathbf{X}^{\dagger}} \left({\mathbf{U}\mathbf{D}\mathbf{V}^\mathrm{T}}\right)} \left({\mathbf{Y}^\mathrm{T}}\right)^{\dagger},
	\end{aligned}
	\label{eq:map}
	\end{equation}
	where $\dagger$ indicates pseudo-inverse.

To keep the uniqueness of rotation, we let $\mathbf{U} = \mathbf{U}^{*}$, $\mathbf{V} = \mathbf{V}^{*}$ to make $\mathbf{R} = \mathbf{R^{*}}$. Then, \equref{eq:map} can be rewritten as
\begin{equation}
	\begin{aligned}
	\mathbf{P} & = {{\mathbf{X}^{\dagger}} \left({\mathbf{U^*}\mathbf{D}\mathbf{V^*}^\mathrm{T}}\right)} \left({\mathbf{Y}^\mathrm{T}}\right)^{\dagger} \\
	& = {{\mathbf{X}^{\dagger}} \left({\mathbf{U^*} \left[ \begin{matrix}
    \sigma_1 &   &   \\
     & \sigma_2 &   \\
     &   & \sigma_3
    \end{matrix} \right] 
      \mathbf{V^*}^\mathrm{T}} \right)} \left({\mathbf{Y}^\mathrm{T}}\right)^{\dagger}.
	\end{aligned}
	\label{eq:map_fixed_uv}
\end{equation}
Based on \equref{eq:map_fixed_uv}, we can draw the conclusion that there are additional degrees of freedoms that are unconstrained in $\mathbf{D}$ even though we fix $\mathbf{U} = \mathbf{U}^{*}$ and $\mathbf{V} = \mathbf{V}^{*}$. In other words, the soft matching matrix $\mathbf{P}$ is not unique even when the rotation matrix is fixed. By giving different values of $\mathbf{D}$, we can achieve different $\mathbf{P}$, but will solve the same rotation matrix. Formally, although a unique rotation matrix $\mathbf{R}^{*} = \mathbf{V^*} \mathbf{U^*}^\mathrm{T}$ could be obtained, a group of $\mathbf{P}$ can be found as
\begin{equation}
	\begin{aligned}
	\mathbf{P} &= g(\mathbf {D}|\mathbf{U^*},\mathbf{V^*}),
	\end{aligned}
	\label{eq:general_form_p}
	\end{equation}
	where $g(\cdot)$ represents the computation of \equref{eq:map_fixed_uv}. $\hfill \blacksquare$

\begin{figure}[!t]
	\centerline{\includegraphics[width=8cm]{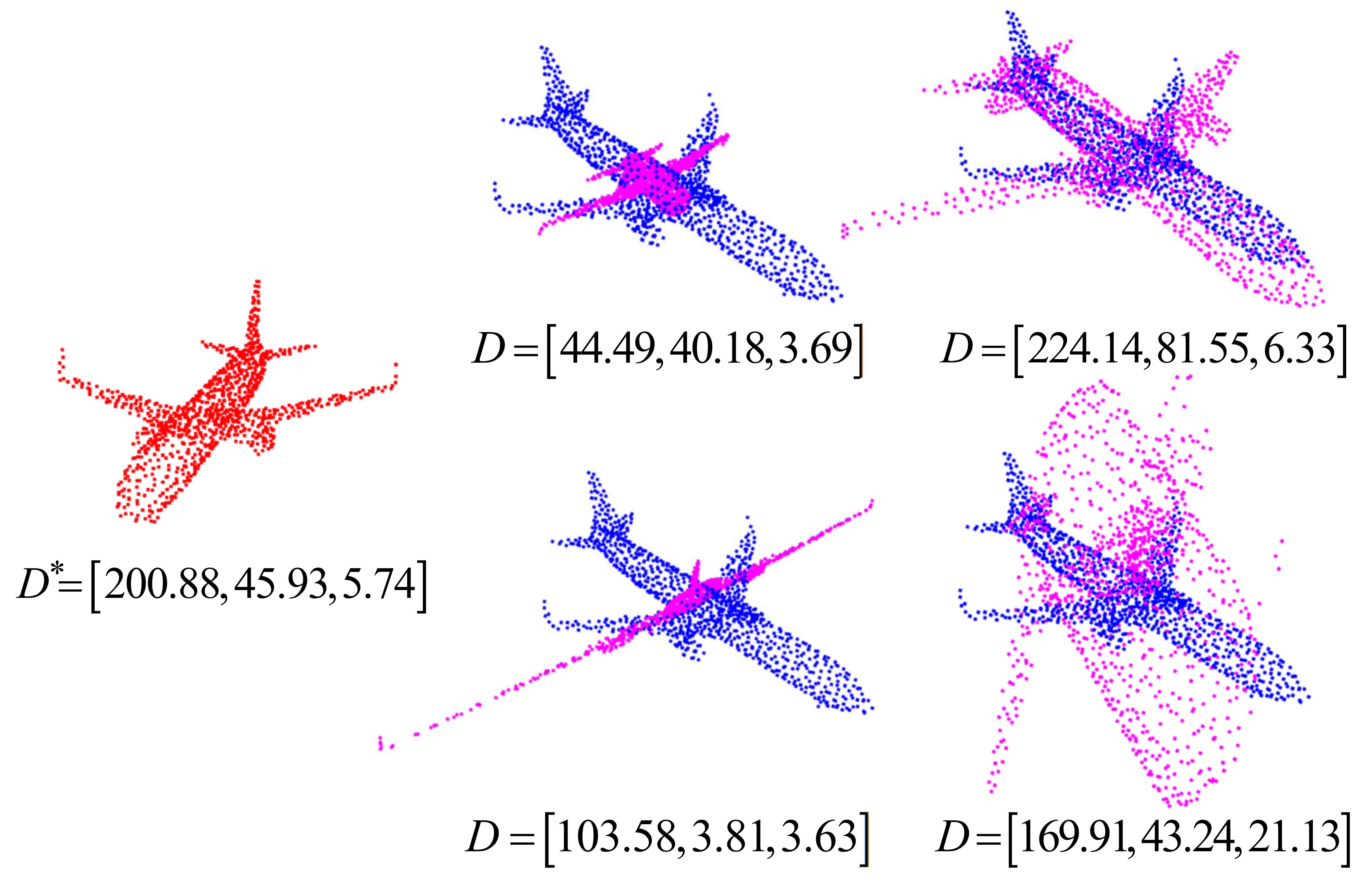}}
	\caption{The point clouds in red and blue are the source and target respectively. The four point sets in pink are constructed virtual points via $\mathcal{Y}\mathbf{P}^\mathrm{T}$ using different soft matching matrices, which are generated by \equref{eq:map_fixed_uv} with different diagonal matrices $\mathbf{D}$. We can easily found that the geometry of these virtual corresponding points has been totally lost. 
	}
	\label{fig:ambigiuty}
\end{figure}

\noindent\textbf{Analysis:} When the ambiguity exists in the consistent subsets $\mathbf{X}$ and $\mathbf{Y}$, \ie there are many virtual point sets corresponding to the same transformation resulted from the Procrustes algorithm \cite{gower_procrustes_1975}. Since $\mathbf{X}$ and $\mathbf{Y}$ are subsets of the input point clouds $\mathcal{X}$ and $\mathcal{Y}$, we can solve the unique rigid motion parameters from different virtual corresponding point sets and $\mathcal{Y}$, \ie the ambiguity of soft matching-based methods exists universally whether the input point clouds are consistent or not. 
This inherent ambiguity will cause the serious distribution degeneration of the virtual corresponding points, which is a special deformation violating the initial assumption of rigid motion. Thus, it cannot be supported by the Procrustes algorithm essentially. Furthermore, this ambiguity will confuse the network, and the learned correspondences are disabled for many downstream tasks.

\noindent\textbf{Examples:} Here, we provide some specific examples of the degeneration due to different $\mathbf{D}$. As shown in \figref{fig:ambigiuty}, a point cloud with 1024 points has been selected as the source (red) from the ``airplane'' category of the ModelNet40 \cite{wu_modelnet40_cvpr_2015} dataset, and then a rotation matrix and a translation vector are applied to the source for the target point cloud (blue). Note that this pair of point clouds is in the same order. By performing the Procrustes algorithm on them, the rigid transformation  $\mathbf{R}^{*}$, $\mathbf{t}^{*}$ are solved, the correct diagonal matrix is obtained as $\mathbf{D^{*}}=[200.88,45.93,5.74]$.
Meanwhile, four different virtual corresponding point sets (pink) are also constructed based on the soft matching strategy via $\mathcal{Y}\mathbf{P}^\mathrm{T}$. Similar to \equref{eq:map_fixed_uv}, we keep $\mathbf{U}$ and $\mathbf{V}$ the same as the correct ones, and only change the diagonal matrix $\mathbf{D}$ for four different soft matching matrices. From \figref{fig:ambigiuty}, we find that serious degeneration exists in these virtual corresponding points. 

In summary, the above proof, analysis, and examples clearly explain why the soft matching-based methods can still obtain acceptable transformation estimation based on the degenerated virtual points. However, this ambiguity will cause unreliable correspondences, which not only limits the registration performance but also hinders downstream applications.

\section{Details of network architecture \\ and implementation}
\label{sec:supp:network}
In this section, we present a detailed introduction of our network architectures, which are achieved by integrating the S2H matching procedure with representative point cloud registration frameworks including DCP \cite{wang_dcp_iccv_2019}, RPMNet \cite{yew_rpmnet_cvpr_2020}, and DGR \cite{choy_dgr_cvpr_2020} notated as $\text{SHM}_\text{DCP}$, $\text{SHM}_\text{RPMNet}$ and $\text{SHM}_\text{DGR}$ respectively. Please refer to original papers of DCP, RPMNet, and DGR for more details. \textit{Our code will be made public}.

\subsection{$\text{SHM}_\text{DCP}$} \label{sec:sup:network1}
\begin{figure*}[!ht]
	\centerline{\includegraphics[width=0.7\linewidth]{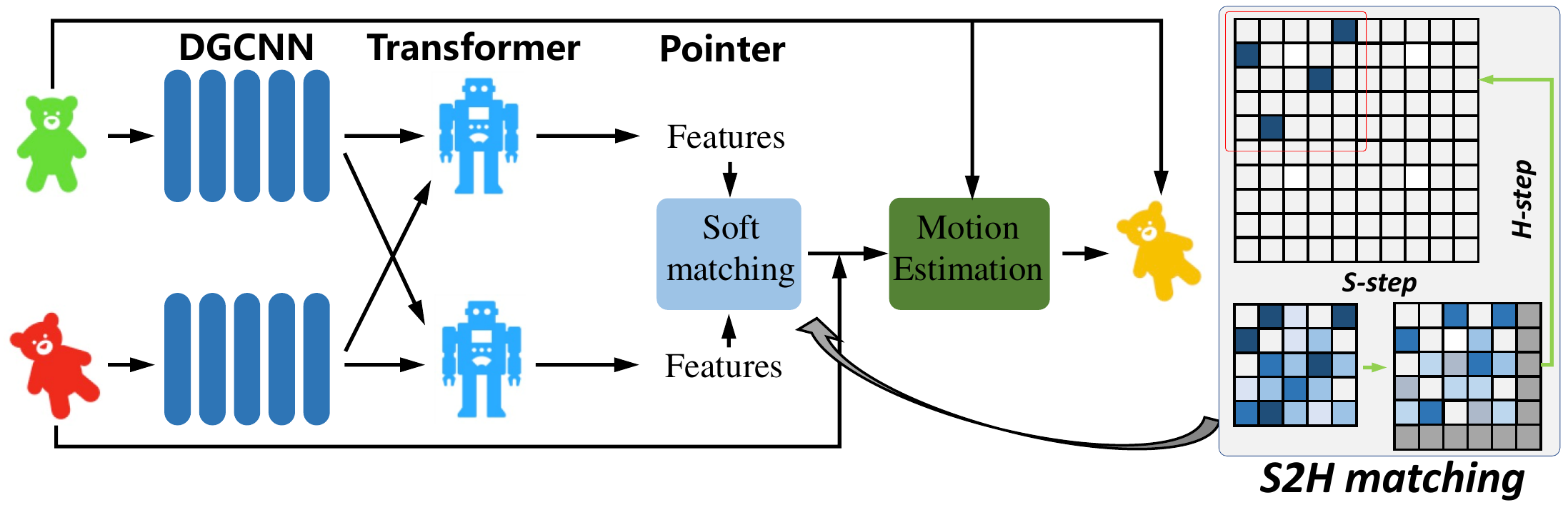}}
	\caption{Illustration of $\text{SHM}_\text{DCP}$ pipeline. Given the source and target, the shared DGCNN and Transformer are employed to extract the point features. Then, via our proposed S2H matching procedure, which is used to replace the soft matching module in the original DCP framework, the PPM is learned to build the correspondences. Finally, the rigid motion is solved by the weighted Procrustes algorithm. $\text{SHM}_\text{DCP}$ is supervised by the complete loss function $\mathcal{L}$.}
	\label{Fig:supp:dcp}
\end{figure*}

The architecture of $\text{SHM}_\text{DCP}$ is presented in \figref{Fig:supp:dcp}, which is achieved by replacing the original matching module with our S2H matching procedure in the DCP framework. The weighted Procrustes algorithm is also applied. Note that the matching module and motion estimation have been introduced in the main paper, we present the feature extraction network here.

We first use DGCNN \cite{wang_dgcnn_tog_2019} to embed point clouds. In DGCNN, we use five edge convolution layers. The numbers of filters of each layer are set to $[64, 64, 128, 256, 512]$ respectively. In each edge convolution layer, BatchNorm is used with the momentum of 0.1, followed by ReLU. Then, we concatenate the outputs from the first four layers and feed them into the last one. Finally, max-pooling is used as our aggregation function for point features.

Then, in the Transformer module \cite{vaswani_attention_nips_2017}, one encoder and one decoder are used, where the multi-head attention with 4 heads is employed. The embedding dimension is set to 512. And a MLP with 1024 hidden dimensions is followed, where ReLU is also used. Inside the Transformer, LayerNorm is used after the multi-head attention and MLP, and before the residual connection. Unlike \cite{vaswani_attention_nips_2017}, we do not use Dropout.

The experiment about $\text{SHM}_\text{DCP}$ is conducted by PyTorch 1.0.0 with CUDA 11.0. The Adam optimizer with an initial learning rate of $1e^{-3}$ is used for training. We train the network for 400 epochs. The complete loss function $\mathcal{L}$ is used here.

\subsection{$\text{SHM}_\text{RPMNet}$} \label{sec:sup:network2}
\begin{figure*}[!ht]
	\centerline{\includegraphics[width=0.7\linewidth]{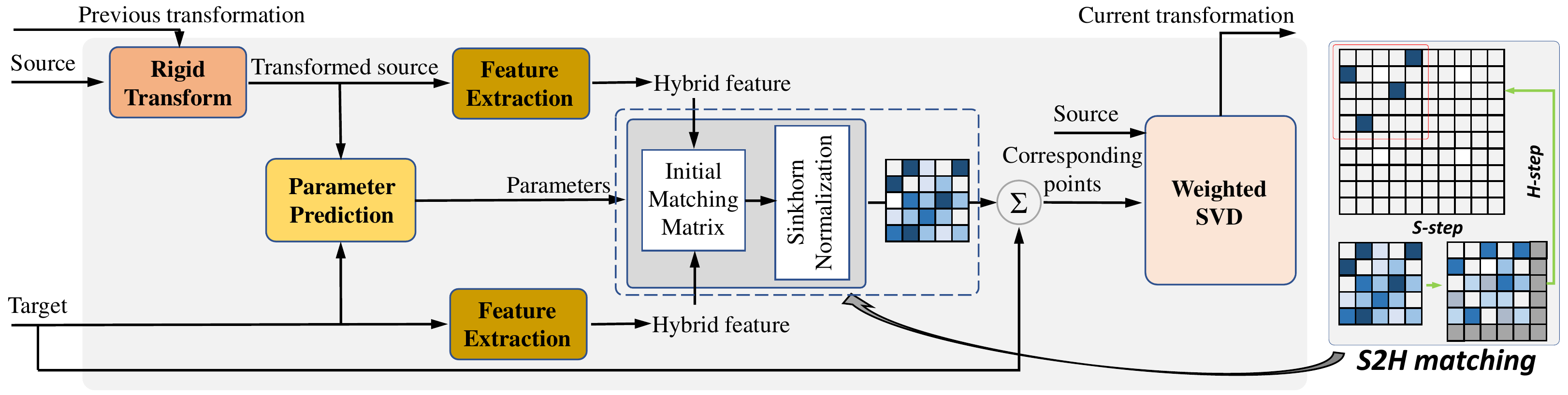}}
	\caption{Illustration of $\text{SHM}_\text{RPMNet}$ pipeline. The iteration strategy is applied in this framework. In each iteration, the hybrid feature is achieved by feeding the 10D hand-craft feature to a PointNet-based network. And then, the PPM is obtained by using the proposed S2H matching procedure instead of the original Augmented-Sinkhorn algorithm. Finally, the weighted Procrustes algorithm is applied to solve the rigid motion. $\text{SHM}_\text{RPMNet}$ is supervised by the complete loss function $\mathcal{L}$.}
	\label{Fig:supp:rpmnet}
\end{figure*}

The architecture of $\text{SHM}_\text{RPMNet}$ is presented in \figref{Fig:supp:rpmnet}, which inherits the framework of RPMNet. Specifically, in addition to the S2H matching procedure, a hybrid feature and the iteration strategy, \ie updating the source by transforming it using the estimated rigid transformation, are both applied in $\text{SHM}_\text{RPMNet}$. 

\begin{table}[!h]
	\renewcommand\arraystretch{1.0}
	\begin{center}
		\resizebox{0.78\linewidth}{!}{
			\begin{tabular}{lcc}
				\toprule
				\textbf{Layer}&Parameters&Output dimension \\
				\midrule				
				\multicolumn{1}{l}{Shared FC}   &$(10,96)$      &$N\times S \times 96$ \\
				\multicolumn{1}{l}{GN + ReLU}   &$(8)$          &$N\times S \times 96$ \\
				\midrule				
				\multicolumn{1}{l}{Shared FC}   &$(10,96)$      &$N\times S \times 96$ \\
				\multicolumn{1}{l}{GN + ReLU}   &$(8)$          &$N\times S \times 96$ \\
				\midrule				
				\multicolumn{1}{l}{Shared FC}   &$(10,96)$      &$N\times S \times 96$ \\
				\multicolumn{1}{l}{GN + ReLU}   &$(8)$          &$N\times S \times 96$ \\
				\midrule
				\multicolumn{1}{l}{Max-pooling}    &$-$            &$N\times 192$ \\
				\midrule				
				\multicolumn{1}{l}{FC}          &$(192,192)$    &$N\times 192$ \\
				\multicolumn{1}{l}{GN + ReLU}   &$(8)$          &$N\times 192$ \\
				\midrule				
				\multicolumn{1}{l}{FC}          &$(192,96)$     &$N\times 96$ \\
				\multicolumn{1}{l}{GN + ReLU}   &$(8)$          &$N\times 96$ \\
				\midrule				
				\multicolumn{1}{l}{FC}          &$(96,96)$      &$N\times 96$ \\
				\multicolumn{1}{l}{$\ell^{2}$ normalization}   &$-$          &$N\times 96$ \\
				\bottomrule
		\end{tabular}}
		\caption{Architecture of feature extraction network in $\text{SHM}_\text{RPMNet}$. It takes a 10D raw feature for each point as input and outputs a 96D feature descriptor.}
		\label{Tab:supp:rpm}
	\end{center}
\end{table}

\begin{figure*}[!ht]
	\centerline{\includegraphics[width=0.7\linewidth]{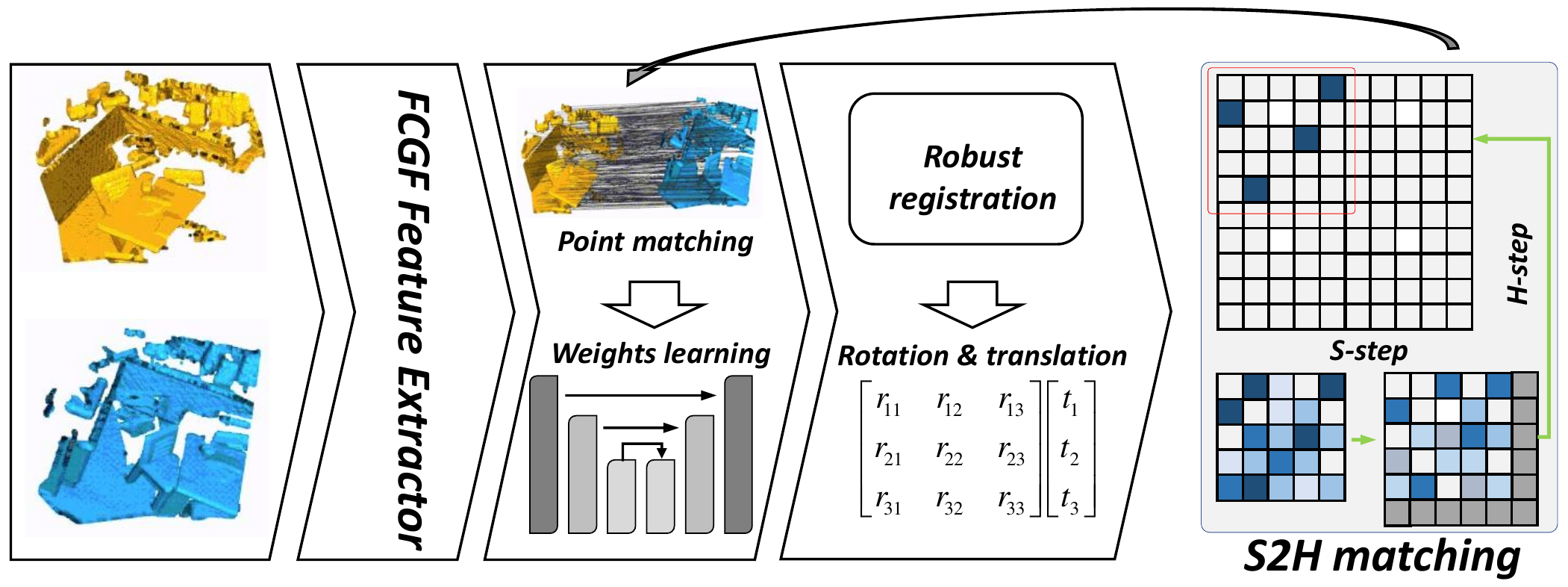}}
	\caption{Illustration of $\text{SHM}_\text{DGR}$ pipeline. FCGF feature extractor is used here to extract point features of large-scale input point clouds. Then, the proposed S2H matching procedure is applied to build the one-to-one correspondences. These constructed correspondences are fed to the weights learning module to select more reliable correspondences furthermore. Finally, the final rigid motion is solved via the robust registration module. It is worth mentioning that, following the protocols of DGR, the feature extractor network has been frozen, and only the weights learning module is trained. Thus, only the rigid motion loss function $\mathcal{L}_3$ is employed here.}
	\label{Fig:supp:dgr} 
\end{figure*}

This hybrid feature is achieved by feeding a 10D hand-crafted feature to a PointNet-based network. 
10D hand-crafted feature consists of the 3D coordinates of the current point, the difference of 3D coordinates between the current point and the neighbor point, and the point pair feature between the current point and the neighbor point (More details can be seen in \cite{yew_rpmnet_cvpr_2020}).
Then, in this PointNet-based network, we use ReLU with group normalization on all layers except the last one. The neighborhood of radius $rad=0.3$ is taken into consideration in feature extraction, and the dimension of the output feature is set to 96.

The detailed structure of this PointNet-based network is presented in \tabref{Tab:supp:rpm}, where $N$ denotes the size of the input point cloud, and $S$ denotes the number of points sampled from the neighborhood cluster for each point. Parameters for fully connected (FC) layers are given as (input-dim, output-dim), and the parameter for group normalization (GN) is the number of groups used.

In addition, we run 2 iterations during the training and use 5 iterations during the test, the results of the final iteration are reported. The experiment is conducted by PyTorch 1.0.0 with CUDA 11.0. We train the network for 1000 epochs using the Adam optimizer with an initial learning rate of $1e^{-4}$. The complete loss function $\mathcal{L}$ is used here.

\subsection{$\text{SHM}_\text{DGR}$} \label{sec:sup:network3}
The architecture of $\text{SHM}_\text{DGR}$ is presented in \figref{Fig:supp:dgr}, which inherits the framework of DGR. Here, we take the FCGF feature \cite{choy_fcgf_iccv_2019} as our feature extractor and a weight learning module is also applied. It is worth mentioning that, following the DGR, this feature extractor network has been frozen, and only the weights learning network is trained. Thus, only the rigid motion loss function $\mathcal{L}_3$ is employed. 
The experiment is conducted by PyTorch 1.5.0 with CUDA 11.0.
We train the network for 100 epochs using the Adam optimizer with an initial learning rate of $1e^{-3}$.

The experiments evaluated on 3DMatch and KITTI totally follow the DGR. Here we provide some details. 
Notably, in Table 5 and Table 6 in the main body, the final reported average RE and TE are computed only on the successfully registered pairs since the relative poses returned from the failed registrations can be drastically different from the ground truth, making the final evaluation results unreliable. 
The input pairs of KITTI are created that are at least 10m apart. And the ground truth transformation is generated by using GPS followed by ICP to fix errors. The input point clouds have been downsampled using the voxel size of 30cm.

\section{Ablation studies}

\subsection{Partial-view and random sampling strategy}
\label{sec::partial_view_and_random_sample}
As stated before, we use the random sampling strategy to generate outliers.
However, a potential risk is that the outliers uniformly distribute. Thus, in this part, we test our S2H matching procedure with a more challenging outliers generation strategy, \ie partial-view \& random sampling.
The experiments are conducted on ModelNet40 in \textit{clean}. 
Specifically, we randomly sample 896 points from the original consistent point clouds, then set a viewpoint in 3D space at an arbitrary position, and select 768 nearest points respectively following PRNet \cite{wang_prnet_nips_2019}. 
We report the results of $\text{SHM}_\text{DCP}$, $\text{SHM}_\text{RPMNet}$ in \tabref{Tab:sample}. We observe that the results are consistent with the random sampling strategy and our S2H matching method helps achieve the best performance under all evaluation metrics.

\begin{table}[h]
	\renewcommand\arraystretch{1.0}
	\begin{center}
		\resizebox{0.95\linewidth}{!}{
			\begin{tabular}{lcccc}
				\toprule
				\textbf{Methods}&$\mathbf{RMSE(R)}$ $\downarrow$&$\mathbf{MAE(R)}$ $\downarrow$&$\mathbf{RMSE(t)}$ $\downarrow$&$\mathbf{MAE(t)}$ $\downarrow$ \\
				\midrule				
				\multicolumn{1}{l}{\text{DCP-v2}}   
				&\multicolumn{1}{c}{6.383}      &\multicolumn{1}{c}{3.984}   &\multicolumn{1}{c}{0.022}       &\multicolumn{1}{c}{0.017}\\
				
				\multicolumn{1}{l}{\text{PRNet}}   
				&\multicolumn{1}{c}{3.806}       &\multicolumn{1}{c}{1.844}    &\multicolumn{1}{c}{0.015}       &\multicolumn{1}{c}{0.011}\\
				
				\multicolumn{1}{l}{$\text{SHM}_\text{DCP}$}          
				&\multicolumn{1}{c}{\textbf{2.805}}  &\multicolumn{1}{c}{\textbf{0.908}} &\multicolumn{1}{c}{\textbf{0.013}} &\multicolumn{1}{c}{\textbf{0.009}}\\
				\midrule
				\multicolumn{1}{l}{\text{RPMNet}}   
				&\multicolumn{1}{c}{1.272}       &\multicolumn{1}{c}{0.527}    &\multicolumn{1}{c}{0.011}       &\multicolumn{1}{c}{0.009}\\
				
				\multicolumn{1}{l}{$\text{SHM}_\text{RPMNet}$}          
				&\multicolumn{1}{c}{\textbf{0.903}}  &\multicolumn{1}{c}{\textbf{0.324}} &\multicolumn{1}{c}{\textbf{0.009}} &\multicolumn{1}{c}{\textbf{0.008}}\\
				\bottomrule
		\end{tabular}}
		\caption{Performance of partial-view and random sampling strategy. }
		\label{Tab:sample}
	\end{center}
\end{table}

\begin{figure*}[!h]
    \centering
	\subfigure[$\text{SHM}_\text{DCP}$ in \textit{clean} setting]{\includegraphics[width=0.45\linewidth]{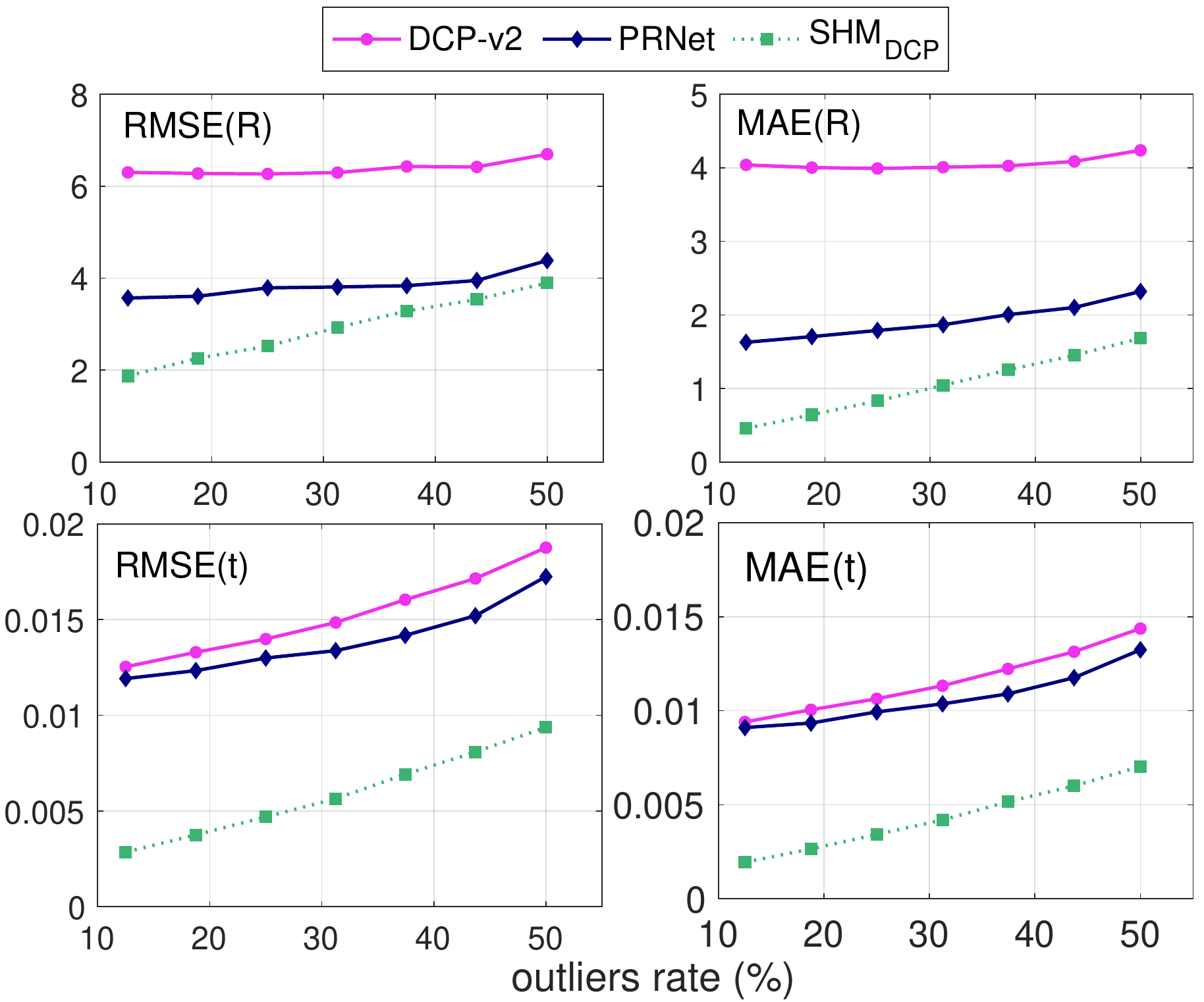}} \quad
	\subfigure[$\text{SHM}_\text{DCP}$ in \textit{noisy} setting]{\includegraphics[width=0.45\linewidth]{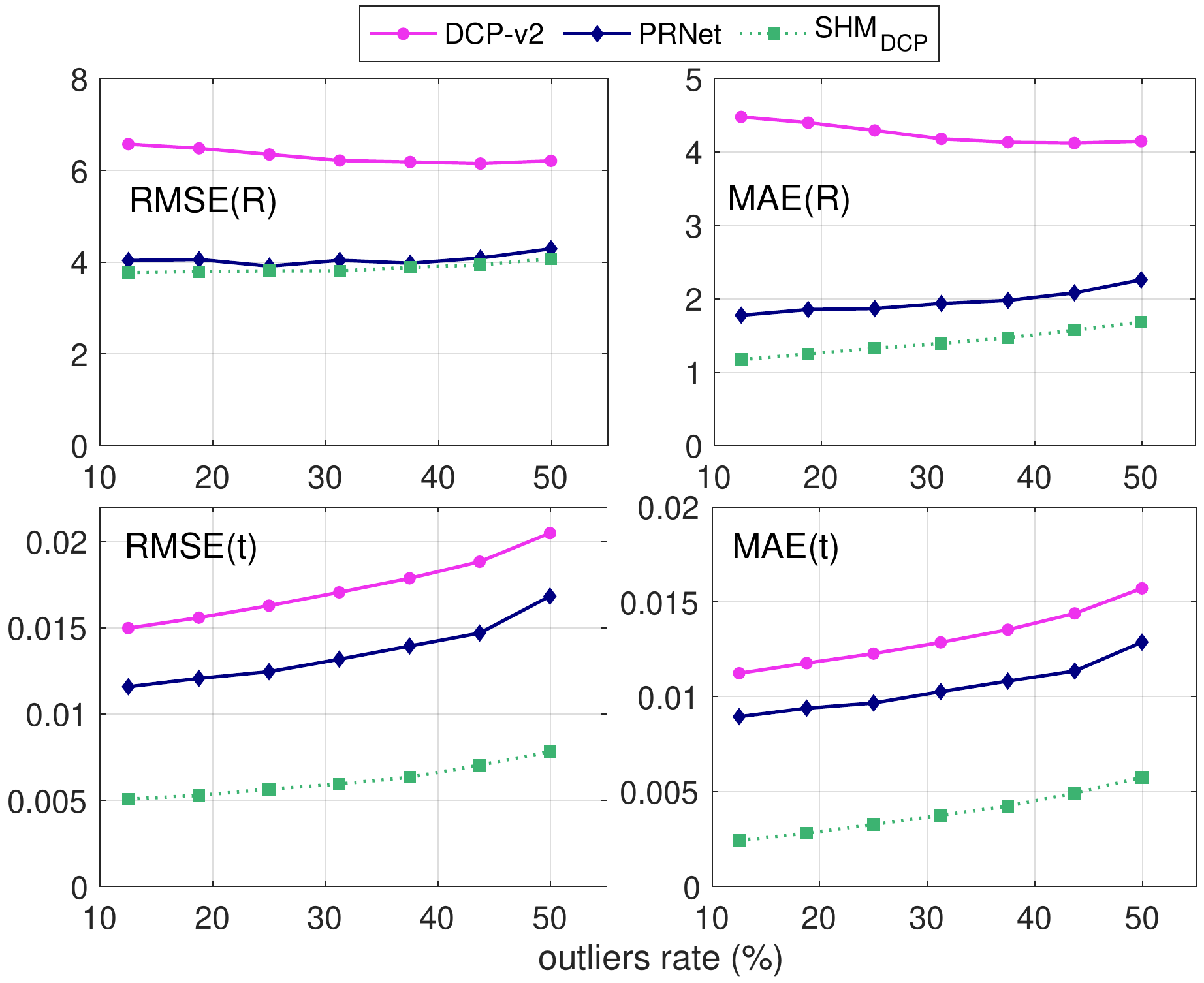}}
	\caption{{Illustration of robustness to outliers of $\text{SHM}_\text{DCP}$.} (a) $\text{SHM}_\text{DCP}$ is better than DCP-v2 and PRNet under different outliers ratios in \textit{clean} setting. (b) $\text{SHM}_\text{DCP}$ is sightly better than PRNet in RMSE(R) and achieves significantly improved performance in MAE(R), RMSE(t) and MAE(t) in \textit{noisy} setting.}
	\label{dcp}
\end{figure*}

\begin{figure*}[!h]
    \centering
	\subfigure[{$\text{SHM}_\text{RPMNet}$ in \textit{clean} setting}]{\includegraphics[width=0.45\linewidth]{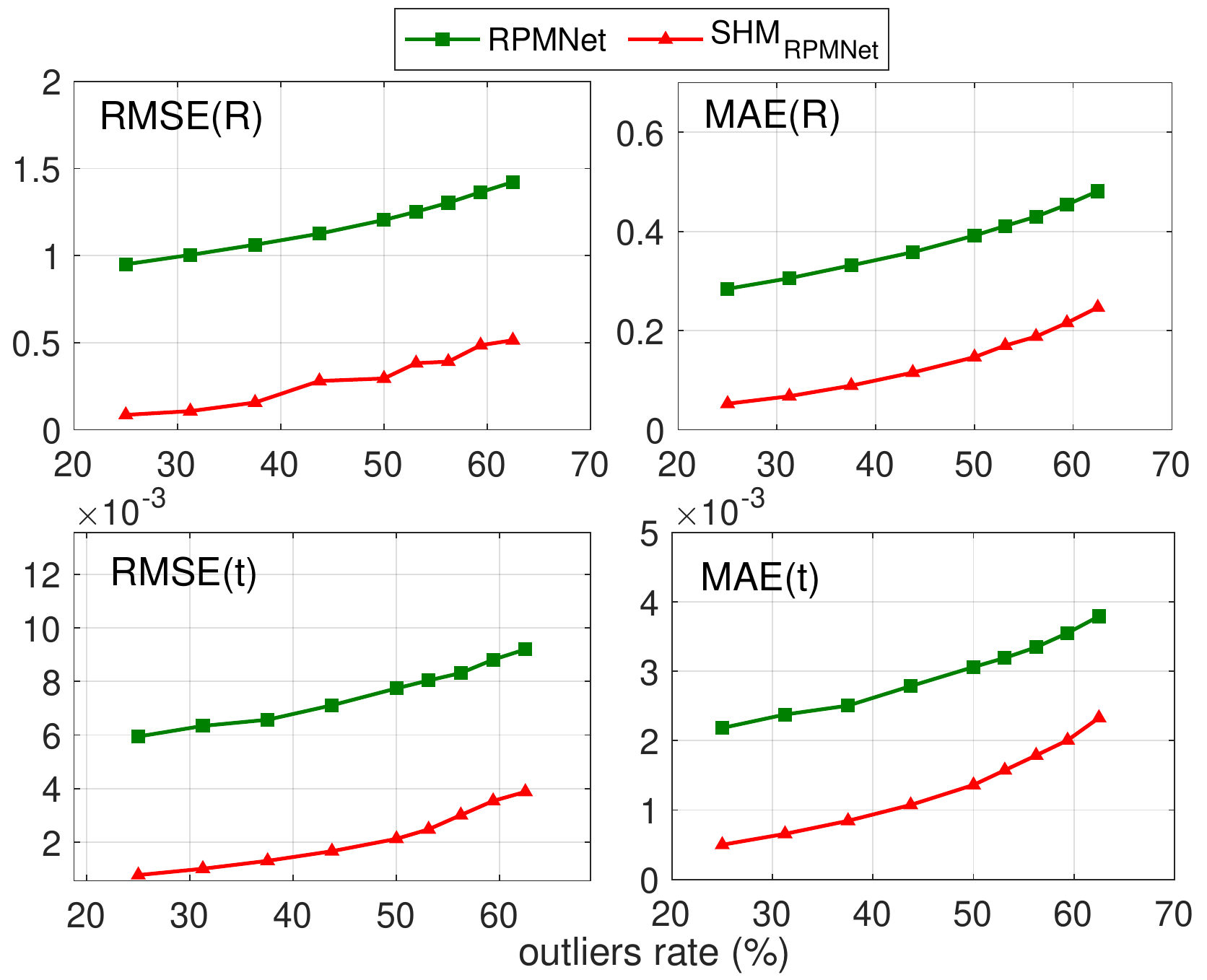}} \quad
	\subfigure[{$\text{SHM}_\text{RPMNet}$ in \textit{noisy} setting}]{\includegraphics[width=0.45\linewidth]{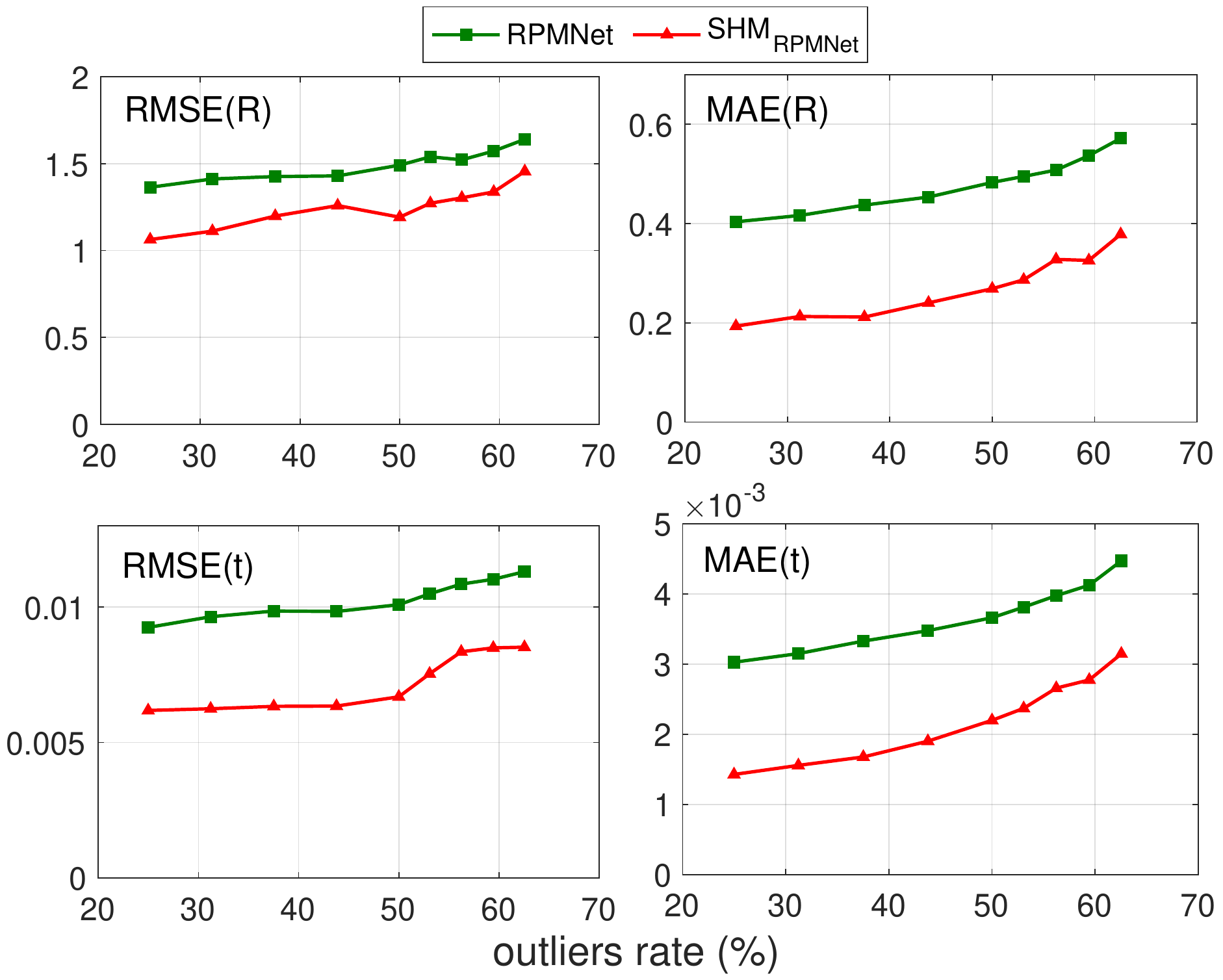}}
	\caption{{Illustration of robustness to outliers of $\text{SHM}_\text{RPMNet}$.} (a) $\text{SHM}_\text{RPMNet}$ is better than RPMNet under different outliers ratios in \textit{clean} setting. (b) In \textit{noisy} setting, $\text{SHM}_\text{RPMNet}$ is also better than RPMNet in all metrics under different outliers ratios.}
	\label{rpm}
\end{figure*}

\subsection{Robustness to Outliers}
\label{sec:robustness_to_outliers}
To verify the robustness to outliers of our proposed method, we evaluate the registration performance of $\text{SHM}_\text{DCP}$ and $\text{SHM}_\text{RPMNet}$ under different outliers ratios since the outliers ratios of the synthetic data can be controlled clearly.

According to \figref{dcp} (a) and \figref{rpm} (a), we can clearly observe that, both $\text{SHM}_\text{DCP}$ and $\text{SHM}_\text{RPMNet}$ obtain better results than the baselines among all evaluation metrics in \textit{clean} under all outliers ratios.
From \figref{dcp} (b) and \figref{rpm} (b), where experiments are conducted in \textit{noisy}, $\text{SHM}_\text{DCP}$ achieves slightly better results than PRNet \cite{wang_prnet_nips_2019} in RMSE(R) metric but significantly advanced performance in MAE(R), RMSE(t), MAE(t) under the outliers ratio of $10\% \sim 50\%$. $\text{SHM}_\text{RPMNet}$ is better than RPMNet in all metrics under the outliers ratio of $25\% \sim 60\%$. These experiment results further validate the robustness of our proposed method.

\subsection{Different loss combinations}
To analyze the effectiveness of the loss function $\mathcal{L}_1$, $\mathcal{L}_2$, and $\mathcal{L}_3$, we conduct evaluations on $\text{SHM}_\text{DCP}$ in \textit{clean} with different loss function combinations herein. The evaluation results are reported in \tabref{Tab:supp:loss}. Using $\mathcal{L}_3$ alone achieves better results than using $\mathcal{L}_1$ alone, we suspect the reason is that too many points are labeled as outliers. And the training does not converge when only $\mathcal{L}_2$ is used. Moreover, when we combine two loss functions to train the network, $\mathcal{L}_1$+$\mathcal{L}_2$ obtains the best performance without the supervision of rigid motion, \ie accurate matching results can even achieve better results than directly supervising the rigid motion. Of cause, the combination of all three loss functions achieves the best performance.

\begin{table}[h]
	\renewcommand\arraystretch{1.0}
	\begin{center}
		\resizebox{0.95\linewidth}{!}{
			\begin{tabular}{lcccc}
				\toprule
				\textbf{Loss}&$\mathbf{RMSE(R)}$ $\downarrow$&$\mathbf{MAE(R)}$ $\downarrow$&$\mathbf{RMSE(t)}$ $\downarrow$&$\mathbf{MAE(t)}$ $\downarrow$ \\
				\midrule				
				\multicolumn{1}{l}{$\mathcal{L}_1$}   
				&\multicolumn{1}{c}{12.472}      &\multicolumn{1}{c}{9.832}   &\multicolumn{1}{c}{0.743}       &\multicolumn{1}{c}{0.697}\\
				\multicolumn{1}{l}{$\mathcal{L}_3$}   
				&\multicolumn{1}{c}{10.322}      &\multicolumn{1}{c}{7.436}   &\multicolumn{1}{c}{0.372}       &\multicolumn{1}{c}{0.216}\\
				\multicolumn{1}{l}{$\mathcal{L}_1$+$\mathcal{L}_2$}   
				&\multicolumn{1}{c}{3.072}       &\multicolumn{1}{c}{1.421}    &\multicolumn{1}{c}{0.017}       &\multicolumn{1}{c}{0.005}\\
				\multicolumn{1}{l}{$\mathcal{L}_1$+$\mathcal{L}_3$}   
				&\multicolumn{1}{c}{5.794}      &\multicolumn{1}{c}{2.834}   &\multicolumn{1}{c}{0.542}       &\multicolumn{1}{c}{0.109}\\
				\multicolumn{1}{l}{$\mathcal{L}_2$+$\mathcal{L}_3$}   
				&\multicolumn{1}{c}{3.283}      &\multicolumn{1}{c}{1.947}   &\multicolumn{1}{c}{0.046}       &\multicolumn{1}{c}{0.037}\\
				\multicolumn{1}{l}{$\mathcal{L}_1$+$\mathcal{L}_2$+$\mathcal{L}_3$}         
				&\multicolumn{1}{c}{\textbf{2.522}}  &\multicolumn{1}{c}{\textbf{0.833}} &\multicolumn{1}{c}{\textbf{0.005}} &\multicolumn{1}{c}{\textbf{0.003}}\\
				\bottomrule
		\end{tabular}}
		\caption{The registration performance of $\text{SHM}_\text{DCP}$ supervised by different loss function combinations. }
		\label{Tab:supp:loss}
	\end{center}
\end{table}

\subsection{More qualitative result}
\begin{figure*}[!h]
	\centerline{\includegraphics[width=0.83\linewidth]{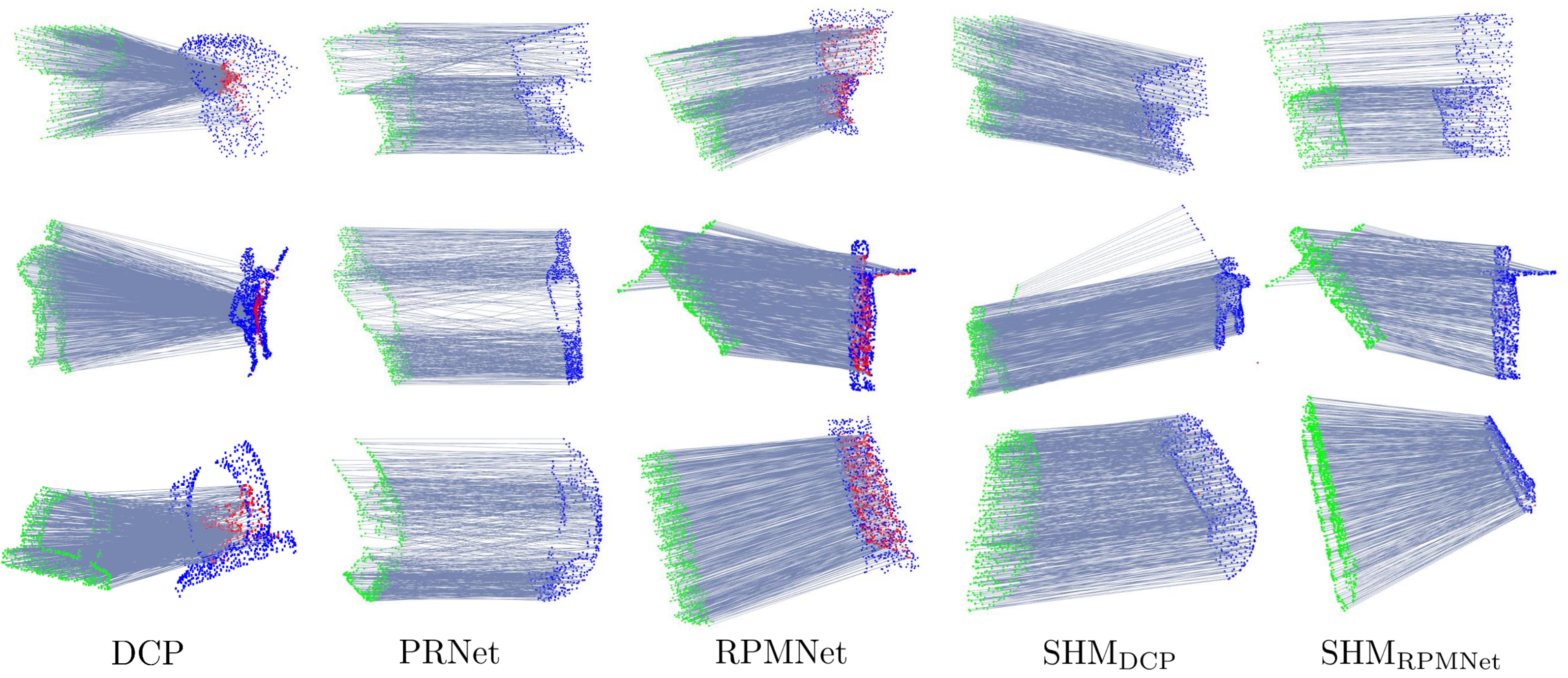}}
	\caption{{The registration results of $\text{SHM}_\text{DCP}$ and $\text{SHM}_\text{RPMNet}$ on ModelNet40}. Different colors indicate the source (green), target (blue) and ``virtual points'' (red) (only in soft matching-based methods, such as DCP and RPMNet) respectively, the connection line indicates the correspondences.}
	\label{Fig:supp:modelnet40}
\end{figure*}

\begin{figure*}[!h]
	\centerline{\includegraphics[width=0.90\linewidth]{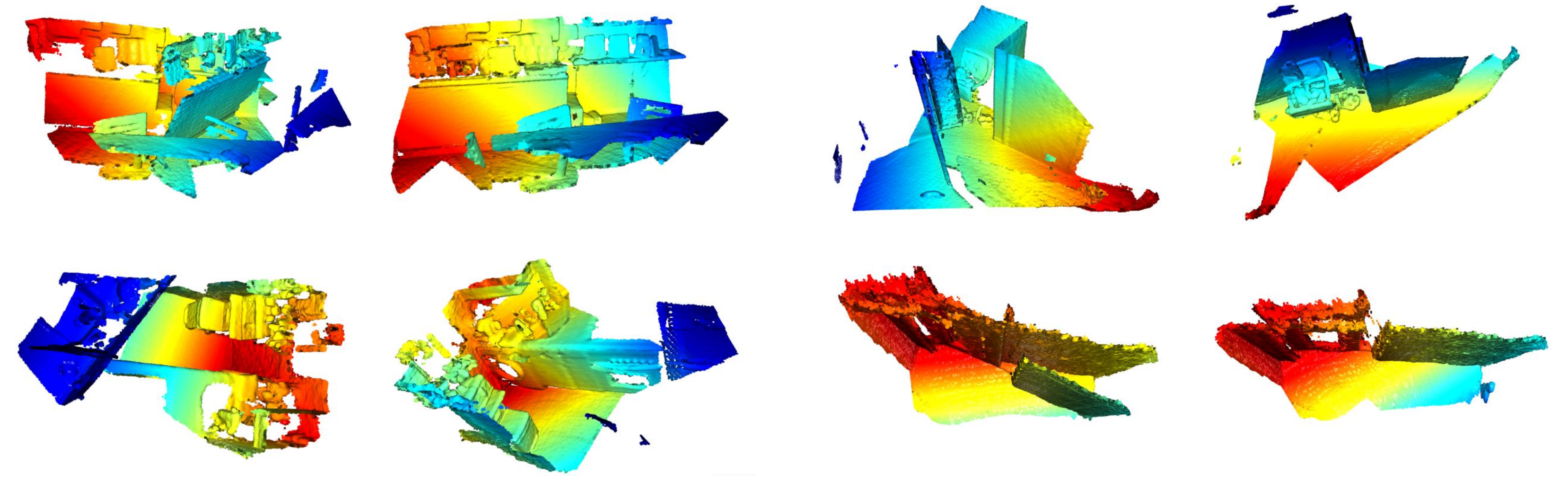}}
	\caption{Registration results of $\text{SHM}_\text{DGR}$ on 3DMatch. For each pair of point clouds, we provide the original input point clouds (\textbf{Left}) and the aligned point clouds \textbf{Right}. }
	\label{Fig:supp:3dmatch}
\end{figure*}

\begin{figure*}[!h]
	\centerline{\includegraphics[width=0.90\linewidth]{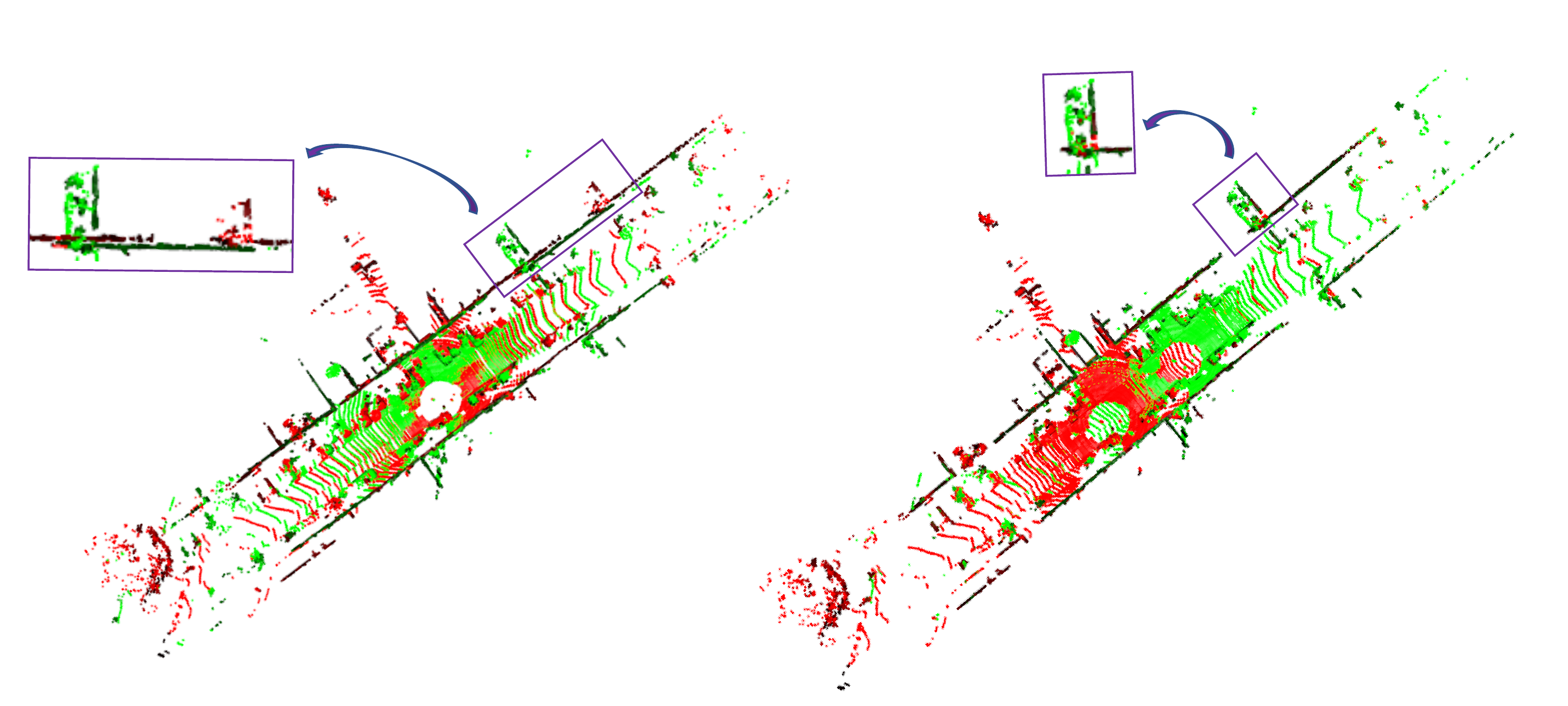}}
	\caption{Registration results of $\text{SHM}_\text{DGR}$ on KITTI. \textbf{Left} is the input point clouds and \textbf{Right} is the aligned point clouds. For a clear comparison, we marked some positions.}
	\label{Fig:supp:kitti}
\end{figure*}

For a more clear understanding of our method, we provide more illustrations of the 3D point cloud registration results on ModelNet40, 3DMatch and KITTI in \figref{Fig:supp:modelnet40}, \figref{Fig:supp:3dmatch} and \figref{Fig:supp:kitti} respectively. Our proposed S2H matching procedure not only helps achieve reliable correspondences avoiding degeneration problem, but also obtains improved registration performance for both synthetic data and real data.

\bibliography{References}